\documentclass[10pt,twocolumn,letterpaper]{article}

\usepackage[pagenumbers]{cvpr} 

\definecolor{cvprblue}{rgb}{0.21,0.49,0.74}
\usepackage[pagebackref,breaklinks,colorlinks,allcolors=cvprblue]{hyperref}
\usepackage{booktabs,multirow,graphicx,array,makecell}
\usepackage{makecell}
\usepackage[table]{xcolor}
\usepackage{float}
\usepackage{pifont}
\usepackage{soul}
\usepackage[ruled,vlined]{algorithm2e}
\definecolor{ltgray}{gray}{0.94}
\definecolor{myblue}{RGB}{37,99,235}   
\renewcommand{\arraystretch}{1.06}
\SetKwInput{KwInit}{Init}
\SetKwInput{KwIn}{Input}
\SetKwInput{KwOut}{Output}
\SetKwFunction{Quantile}{Quantile}
\SetKw{Continue}{continue}
\newcolumntype{M}[1]{>{\centering\arraybackslash}m{#1}}

\usepackage{mathtools}
\makeatletter
\g@addto@macro\normalsize{%
  \setlength\abovedisplayskip{8pt plus 2pt minus 2pt}%
  \setlength\belowdisplayskip{8pt plus 2pt minus 2pt}%
  \setlength\abovedisplayshortskip{6pt plus 2pt minus 2pt}%
  \setlength\belowdisplayshortskip{6pt plus 2pt minus 2pt}%
}
\makeatother
\usepackage{adjustbox}
\def\paperID{****} 
\def\confName{CVPR}
\def\confYear{2026}

\title{Learning through Creation: A Hash-Free Framework for On-the-Fly Category Discovery}

\author{
Bohan Zhang$^{1}$\thanks{These authors contributed equally to this work.},
Weidong Tang$^{1}$\footnotemark[1],
Zhixiang Chi$^{2}$, Yi Jin$^{3}$, \\
Zhenbo Li$^{1}$, Yang Wang$^{4}$, Yanan Wu$^{1}$\thanks{Corresponding author.} \\
$^{1}$College of Information and Electrical Engineering, China Agricultural University, China \\
$^{2}$Department of Electrical and Computer Engineering, University of Toronto, Canada \\
$^{3}$School of Computer and Information Technology, Beijing Jiaotong University, China \\
$^{4}$Department of Computer Science and Software Engineering, Concordia University, Canada \\
{\ttfamily\small zbohan082@gmail.com \quad wdtang29@gmail.com \quad zhxchi@ece.utoronto.ca} \\
{\ttfamily\small    yjin@bjtu.edu.cn \quad \{lizb,ynwu\}@cau.edu.cn \quad yang.wang@concordia.ca}
}

\begin{document}
\maketitle
\begin{abstract}
On-the-Fly Category Discovery (OCD) aims to recognize known classes while simultaneously discovering emerging novel categories during inference, using supervision only from known classes during offline training. Existing approaches rely either on fixed label supervision or on diffusion-based augmentations to enhance the backbone, yet none of them explicitly train the model to perform the discovery task required at test time. It is fundamentally unreasonable to expect a model optimized on limited labeled data to carry out a qualitatively different discovery objective during inference. This mismatch creates a clear optimization misalignment between the offline learning stage and the online discovery stage. In addition, prior methods often depend on hash-based encodings or severe feature compression, which further limits representational capacity. To address these issues, we propose Learning through Creation (LTC), a fully feature-based and hash-free framework that injects novel-category awareness directly into offline learning. At its core is a lightweight, online pseudo-unknown generator driven by kernel-energy minimization and entropy maximization (MKEE). Unlike previous methods that generate synthetic samples once before training, our generator evolves jointly with the model’s dynamics and synthesizes pseudo-novel instances on-the-fly at negligible cost. These samples are incorporated through a dual max-margin objective with adaptive thresholding, strengthening the model’s ability to delineate and detect unknown regions through explicit creation. Extensive experiments across seven benchmarks show that LTC consistently outperforms prior work, achieving improvements ranging from 1.5\% to 13.1\% in all-class accuracy. The code is available at \textcolor{myblue}{https://github.com/brandinzhang/LTC}
\end{abstract}

\section{Introduction}
\label{sec:intro}
\begin{figure}[t]
    \centering
    \includegraphics[width=1\linewidth]{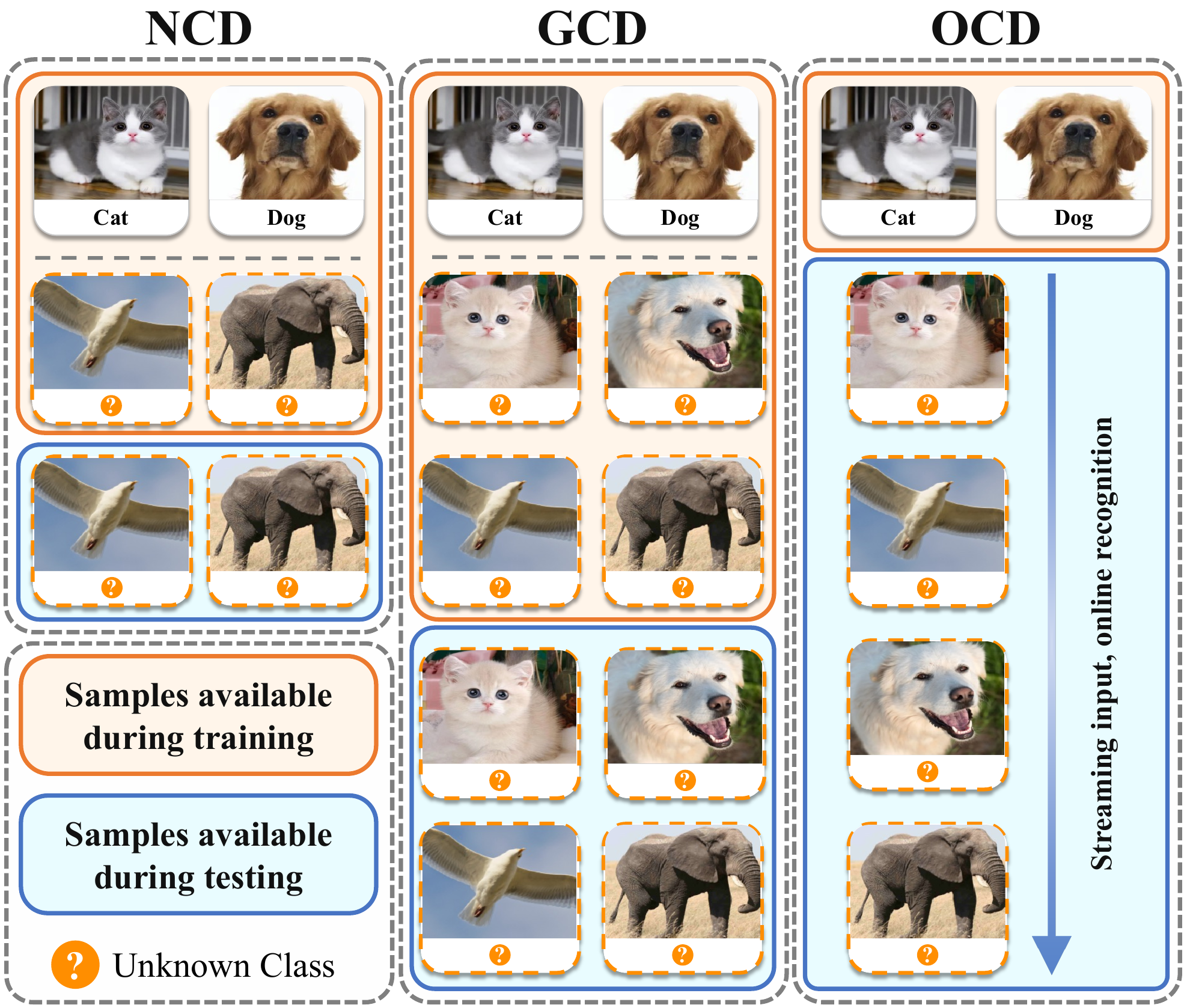}
    \caption{\textbf{Comparison of category discovery settings.}  Unlike NCD/GCD, which rely on a pre-defined query set during training, OCD enables new category discovery directly from streaming data without prior access to test samples.}

    \label{CD_compare}
\end{figure}

Conventional deep learning models are typically trained under a closed-world assumption with a fixed set of categories~\cite{he2015delving,dosovitskiy2020image,he2016deep,krizhevsky2012imagenet,huang2017densely,wu2023metazscil}. Their ability to recognize unseen classes is therefore forfeited, which is unrealistic in dynamic real-world environments~\cite{chi2022metafscil,liu2022few}. The challenge becomes even more pronounced when emerging categories lack annotations~\cite{chi2025plug}. Novel Category Discovery (NCD)~\cite{NCD} and Generalized Category Discovery (GCD)~\cite{2022GCD,rastegar2023learn,zhao2023learning} aim to address this limitation by transferring knowledge from labeled data to uncover novel classes within unlabeled samples, while preserving recognition of known ones. Despite recent advances~\cite{wu2023metagcd}, the majority of these methods still require known and novel classes to be jointly available during training (see Figure~\ref{CD_compare} (NCD and GCD)), leading to repetitive large-scale retraining when new unlabeled data arrive. Their reliance on offline inference also limits applicability in online settings that demand real-time feedback.

To address these limitations, On-the-Fly Category Discovery (OCD)~\cite{SMILE} has been introduced as a more practical and realistic paradigm. It comprises an offline stage, where the model is trained on a labeled dataset, and an online stage, where streaming unlabeled instances arrive sequentially and may belong to either known or novel categories (see Figure~\ref{CD_compare} (OCD)).
Instance-based discovery and recognition in OCD present inherent challenges. The pioneering work SMILE~\cite{SMILE} introduces a hash-based framework in which extracted features are quantized into binary codes for recognition and prototype construction. But this discretization diminishes fine-grained feature cues and limits representational richness. PHE~\cite{PHE} further proposes a prototypical supervision strategy to enhance expressiveness, yet it remains confined within the same hash-based paradigm. 


DiffGRE~\cite{DiffGRE} attempts to move away from hash-based frameworks, yet its projection of rich, high-dimensional features into an extremely narrow 12-dimensional space still discards substantial semantic information. Its generative component, which produces samples for both known and pseudo-novel concepts, ultimately serves only as a data augmentation mechanism rather than a discovery mechanism. This exposes a central flaw shared by all prior methods: they focus almost exclusively on enhancing representation quality while ignoring the core objective of OCD. Expecting a model trained solely for representation refinement to suddenly perform category discovery at inference is fundamentally misguided. In practice, these approaches never expose the model to the discovery task during training, resulting in a clear optimization failure and leaving the model ill-equipped for online category discovery.
 


Our key insight is that category discovery must be trained in a \textit{discovery-oriented} manner that mirrors the evaluation process~\cite{chi2021test,wu2024test}. In OCD, the model has no access to unknown categories during training, which makes it unrealistic to expect discovery to emerge without explicit guidance. A model that never learns what “unknown” looks like cannot be expected to detect it at test time. To address this fundamental oversight, we introduce Learning through Creation (LTC), a framework that actively generates and integrates pseudo-novel samples during training. These created samples provide direct feedback at every iteration, shaping the boundary between known and unknown regions and aligning optimization with the actual inference objective.

Unlike DiffGRE, which pre-generates synthetic data once and relies entirely on heavy pretrained diffusion models~\cite{rombach2022high}, LTC performs generation online and adapts continually to the evolving model. At each step, a single-step entropy–kernel objective produces off-support samples using only the current features and the current model state. These samples are then utilized through a dual max-margin loss with an adaptive threshold, enforcing robust representation learning and explicit unknown discovery. In parallel, LTC eliminates hash encodings and adopts continuous prototype matching, avoiding discretization artifacts and improving stability throughout training. Extensive experiments confirm that LTC delivers substantial improvements over state-of-the-art approaches across all benchmarks.
Our main contributions are summarized as follows:

\begin{itemize}
\item We propose Learning through Creation, a novel training strategy that explicitly learns the discovery objective, addressing a key limitation of previous methods that expect discovery without directly optimizing for it.
\item We design an adaptive pseudo-sample generator that produces semantically meaningful samples beyond decision boundaries. Combined with a dual max-margin loss and adaptive thresholding, it establishes strong known–unknown separation that prior approaches fail to capture.
\item We introduce a hash-free framework that replaces discrete hash codes with continuous prototype embedding, eliminating discretization artifacts and enabling stable, expressive on-the-fly category expansion.
\item Extensive experiments on seven OCD benchmarks demonstrate the superiority of our LTC framework, achieving 1.5\%–13.1\% improvements in all-class accuracy over state-of-the-art methods.
\end{itemize}





\section{Related Work}
\label{sec:formatting}

\noindent\textbf{Category Discovery.}
Novel Category Discovery (NCD), first explored by DTC~\cite{NCD}, transfers supervision from labeled known classes to organize unlabeled data into novel categories. Early approaches~\cite{NCD,fini2021unified,wang2024semantic,zhao2021novel,wang2021progressive,zang2023boosting} typically assume that all unlabeled samples belong to unseen classes. Generalized Category Discovery (GCD)~\cite{2022GCD} relaxes this assumption by allowing the unlabeled pool to mix known and novel categories. Despite strong results of subsequent NCD/GCD methods~\cite{2022GCD,wang2025get,pu2023dynamic,wu2023metagcd,wen2023parametric,yang2022divide,zhao2023incremental,an2023generalized,ma2024active}, two standard assumptions limit practical use: (i) training relies on a pre-defined query set of unlabeled data, which can bias models toward that set and weaken generalization to truly new inputs; and (ii) inference is performed offline in batches, misaligned with dynamic settings. To address these issues, Du \emph{et al.} introduced On-the-Fly Category Discovery (OCD)~\cite{SMILE}, removing the fixed-query-set requirement and enabling instance-level feedback for streaming inputs; their method assigns categories via the sign pattern of learned representations (hash-form codes). Zheng \emph{et al.} further proposed PHE~\cite{PHE}, representing each category with multiple hash prototypes and adopting a finer novel-class decision rule, yielding improved accuracy. \emph{These existing pipelines all compress features to the hash space through a hash head, which may discard fine-grained cues; we therefore present a hash-free baseline and build additional modules on top for further gains.}

\noindent\textbf{Data Synthetic Methods.}  
Synthetic data generation has long served as an effective tool for improving model generalization. Classic techniques such as Mixup~\cite{zhang2017mixup} and CutMix~\cite{yun2019cutmix} augment training by linearly or regionally mixing samples from known classes. More recently, methods like DiffuseMix~\cite{wang2024enhance} leverage pretrained diffusion models to produce semantically consistent mixed samples, further enriching the known-class training distribution. While these approaches have proven beneficial, they are inherently limited to synthesizing data \emph{within} the known-class manifold. In contrast, our goal is to synthesize samples that resemble \emph{unknown classes} to simulate OCD scenarios.

Recent works have explored synthesizing unknown-like samples~\cite{li2025generalized,DiffGRE,huang2025open,jang2024synthetic}, but they are not suitable for OCD.
DiffGRE~\cite{DiffGRE} uses a diffusion-based ACG module to generate a large pool of virtual images, filtered by similarity thresholds and matched via the Hungarian algorithm. These samples receive known or virtual labels based on cluster composition. However, the goal is to enrich training data and enhance representations \emph{offline}, rather than synthesize unknowns as a direct training objective under OCD constraints. GCDUSG~\cite{li2025generalized} targets GCD, where access to unlabeled unknowns reduces the need for generation.
Other methods~\cite{huang2025open,jang2024synthetic} rely on offline generators or adversarial distillation and are incompatible with online inference.\emph{Our approach differs in several key ways:}
(i) We avoid the use of an external heavy generator. Instead, we apply one-step entropy–kernel updates from mixup anchors, using only the current model and batch features, which aligns with the “no unknowns at training time” constraint; 
(ii) Our method is online and lightweight. Generation is triggered per batch, maintaining only a small set of pseudo novel class samples per iteration, rather than relying on a large synthetic pool; 
(iii) Our approach is both fair and effective. The performance gains come from shaping the decision boundary rather than increasing the data volume. For example, on \textit{Pets}, generating approximately 200 pseudo novel class samples over the entire training process results in about a 10\% improvement in novel-class discovery, using far fewer synthetic samples compared to augmentation-heavy baselines.


\section{Methodology}

\noindent\textbf{Problem definition.}
On-the-fly category discovery (OCD) aims to continuously recognize known categories while discovering novel ones from a streaming sequence of unlabeled data, using only prior knowledge from labeled known classes. Formally, we are provided with a labeled support set $\mathcal{D}_S = \{(\mathbf{x}_i, y_i^s)\}_{i=1}^M \subseteq \mathcal{X} \times \mathcal{Y}_S$ for training, and an unlabeled query set $\mathcal{D}_Q = \{(\mathbf{x}_j, y_j^q)\}_{j=1}^N \subseteq \mathcal{X} \times \mathcal{Y}_Q$ for evaluation, where $\mathcal{Y}_S \subseteq \mathcal{Y}_Q$.
The label spaces $\mathcal{Y}_S$ and $\mathcal{Y}_Q \setminus \mathcal{Y}_S$ correspond to known/old and novel/unknown categories, respectively.
Only $\mathcal{D}_S$ is accessible during training. At test time, data in $\mathcal{D}_Q$ arrive sequentially in a \emph{streaming} manner, with each sample processed online as it appears.
This continual, instance-level inference paradigm makes OCD fundamentally distinct from conventional closed-world recognition and offline category discovery tasks.


\begin{figure*}[t]
  \centering
  \includegraphics[width=0.95\textwidth]{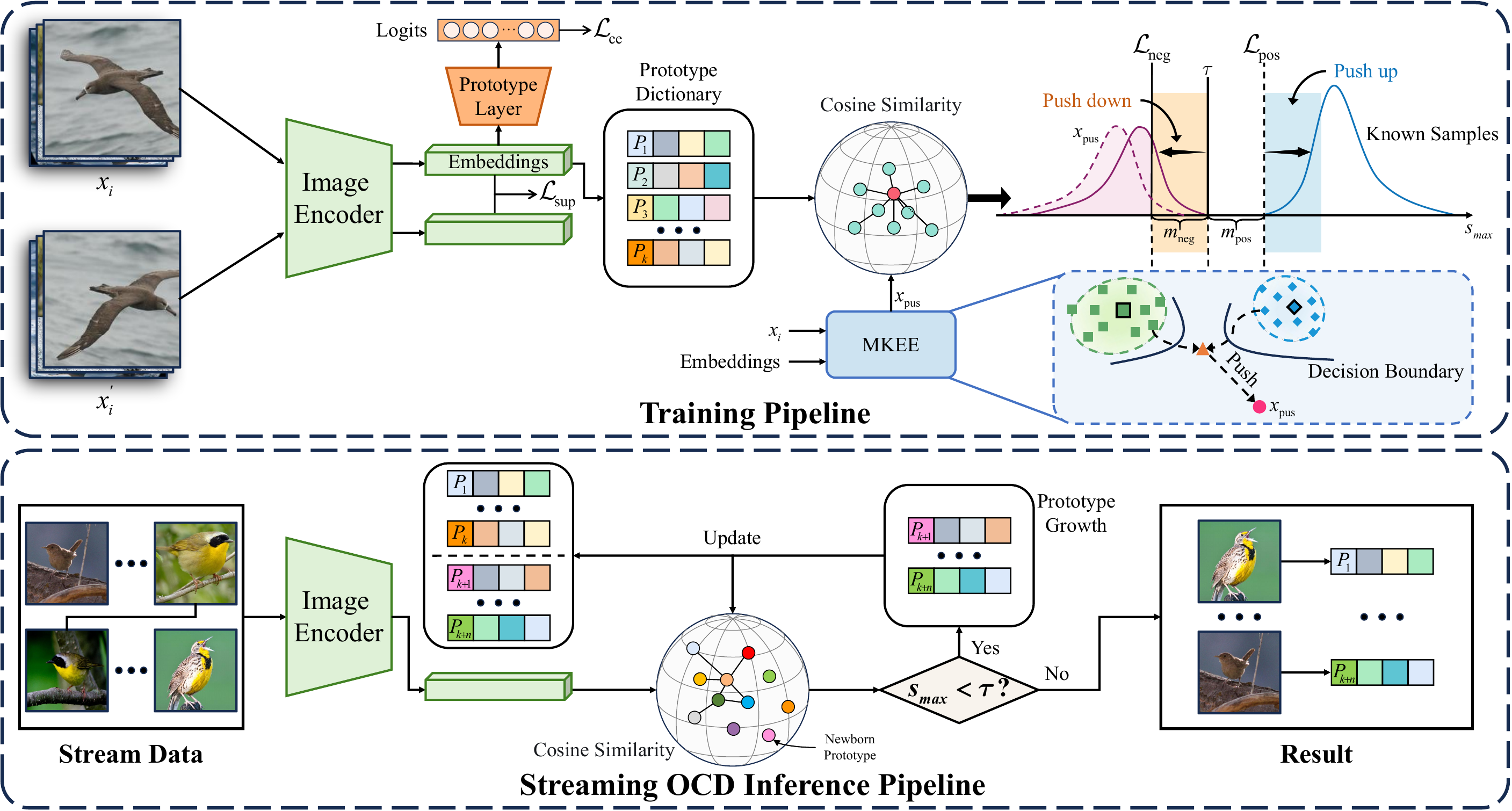}
  \caption{\textbf{Overview of the proposed LTC framework.} During training, we extract feature representations from known-class samples and generate pseudo-unknowns via mixup and MKEE perturbation. A dual max-margin loss enforces separation between known and unknown regions, guided by an adaptive threshold. At inference, dynamic prototype matching enables online discovery of novel categories.}
  \label{fig:framework}
\end{figure*}

\subsection{Baseline: Dynamic Prototype Dictionary}\label{subsec:baseline}

A key limitation of existing OCD methods lies in their dependence on hash-based representations, which often discard fine-grained semantics and cause training instability due to approximate binarization.
To address this issue, we propose a purely feature-based framework that emphasizes semantic expressiveness and representation robustness.
We argue that effective features for OCD should meet two essential criteria:
(i) preserve fine-grained intra-class structure to distinguish subtle variations among categories—particularly crucial in OCD, where novel classes frequently differ in nuanced ways, as also noted in~\cite{SMILE}; and
(ii) enforce strong inter-class separability to maintain discriminative prototypes and enable reliable discovery of new categories.
Accordingly, we adopt a joint training objective that integrates supervised contrastive learning with standard cross-entropy classification.

\noindent\textbf{Supervised contrastive loss.}  
Given an input image $x$ and its augmented view $\hat{x}$, the encoder $z(\cdot)$ produces embeddings $z(x)$ and $z(\hat{x})$, which are $\ell_2$-normalized into features $f(x)$ and $f(\hat{x})$.
To preserve fine-grained semantics and promote class-consistent representations, we employ a supervised contrastive objective following~\cite{seifi2024ood,khosla2020supervised}.
These studies show that incorporating label supervision with data augmentations substantially improves fine-grained feature discrimination~\cite{ye2026normalized}.
Let $\mathcal{P}(i)$ denote the set of positive samples for instance $i$ (including its augmentations and same-class examples), and $\mathcal{B}$ be the batch index set.
With a temperature $T$, the supervised contrastive loss is defined as:


\begingroup
\thinmuskip=0mu
\medmuskip=0mu
\thickmuskip=0mu
\begin{equation}
\hspace{-0.5em}
\resizebox{0.92\linewidth}{!}{$
\begin{aligned}
\mathcal{L}_{\text{sup}}^{(i)}
&= - \frac{1}{|\mathcal{P}(i)|}
   \sum_{x_p\in\mathcal{P}(i)}
   \log
   \frac{
     \exp\!\Big(\dfrac{f(x_i)^{\!\top}\!f(x_p)}{T}\Big)
   }{
     \sum\limits_{b\in\mathcal{B}\!\setminus\!\{i\}}\!
       \exp\!\Big(\dfrac{f(x_i)^{\!\top}\!f(x_b)}{T}\Big)
   }.
\end{aligned}
$}
\end{equation}
\endgroup

\noindent\textbf{Cross-entropy loss.}  
To further strengthen prototype discriminability, a linear classification head maps each feature to known-class logits $\ell(x) \in \mathbb{R}^K$.
The standard cross-entropy objective is formulated as:
\begin{equation}
\mathcal{L}_{\text{ce}} = -\frac{1}{|\mathcal{B}|}\sum_{i\in\mathcal{B}} \log \frac{
\exp\!\big(\ell_{y_i}(x_i)\big)
}{
\sum_{k=1}^K \exp\!\big(\ell_{k}(x_i)\big)
}.
\end{equation}

Let $\mathcal{L}_{\text{sup}} = \frac{1}{|\mathcal{B}|} \sum_{i \in \mathcal{B}} \mathcal{L}_{\text{sup}}^{(i)}$ denote the supervised contrastive loss computed over the batch. The overall training objective jointly optimizes the contrastive and cross-entropy losses, with $\alpha$ balancing their contributions:
\begin{equation}
\mathcal{L}_{\text{total}} = \alpha \mathcal{L}_{\text{sup}} + \mathcal{L}_{\text{ce}}
\end{equation}

\noindent\textbf{Inference with a dynamic prototype dictionary.}  
In contrast to hash-based inference that compares discrete hash codes, our method performs classification directly in the continuous feature space through prototype matching.
At test time, we maintain a dynamic prototype dictionary that evolves as new categories emerge.
For each known class $k$, let $N_k$ denote the number of samples labeled as class $k$. Its prototype is defined as the $\ell_2$-normalized mean of its feature vectors, following prior work in prototype learning~\cite{snell2017prototypical}:
\begin{equation}
P_k = \frac{\frac{1}{N_k} \sum_{i:\,y_i=k} f(x_i)}{\left\| \frac{1}{N_k} \sum_{i:\,y_i=k} f(x_i) \right\|_2},
\quad k = 1, \dots, K,
\end{equation}

During inference, a test sample $x_t$ is assigned to the prototype with the highest cosine similarity:
\begin{equation}
\hat{y}(x_t) = \arg\max_k\, f(x_t)^\top P_k.
\end{equation}

If the maximum similarity $\max_k f(x_t)^\top P_k$ falls below a predefined threshold $\tau$, the sample is considered to belong to a novel class.
A new prototype is then initialized as $P_{K+1} \leftarrow f(x_t)$ and appended to the dictionary.
This simple yet effective mechanism supports on-the-fly category expansion while requiring minimal tuning.
Furthermore, as discussed in Sec.~\ref{subsec:autoTau}, we introduce an adaptive thresholding strategy to automatically adjust $\tau$ and enhance robustness to varying data distributions.

\subsection{Generating Pseudo-Novel Classes via MKEE}
\label{subsec:pus}
OCD presents a more challenging and realistic setting than related tasks such as NCD or GCD.
While the latter allow access to unlabeled unknown samples during training, OCD restricts supervision strictly to labeled data from known classes.
As a result, the model must recognize and adapt to novel categories \emph{without prior exposure}, which is especially difficult in fine-grained scenarios where inter-class differences are subtle.
To address this challenge, we propose a simple yet effective strategy to simulate novel class samples during training.
The core idea is that the model should not only learn discriminative features for known categories but also learn to separate them from pseudo-novel ones, aligning the training objective with the test-time goal of discovering unseen classes~\cite{chi2025learning,chi2024adapting}.
Our method, termed \emph{MKEE} (Minimizing Kernel Energy and Maximizing Entropy), leverages the model’s evolving feature space to generate pseudo-novel samples, improving its ability to recognize and adapt to new categories during inference.

\noindent\textbf{Mixup as manifold-conforming initialization.}  
We first employ mixup~\cite{zhang2017mixup} to synthesize candidate pseudo-novel samples:
\begin{equation}
x_{\text{mix}} = \lambda x_i + (1 - \lambda) x_j, \quad \lambda \sim \mathrm{Beta}(\eta, \eta),
\end{equation}
where \(\eta\) controls the interpolation strength.  
Interpolating between samples from different known classes (\(y_i \neq y_j\)) generates synthetic data near inter-class boundaries while preserving object structure and texture.  
Prior studies~\cite{zhang2017mixup, verma2019manifold} have shown that such interpolants remain close to the natural data manifold, providing a meaningful initialization for generating pseudo-novel examples.

\begin{figure}[!t]
    \centering
    \begin{minipage}[t]{0.495\linewidth}
        \centering
        \includegraphics[width=\linewidth]{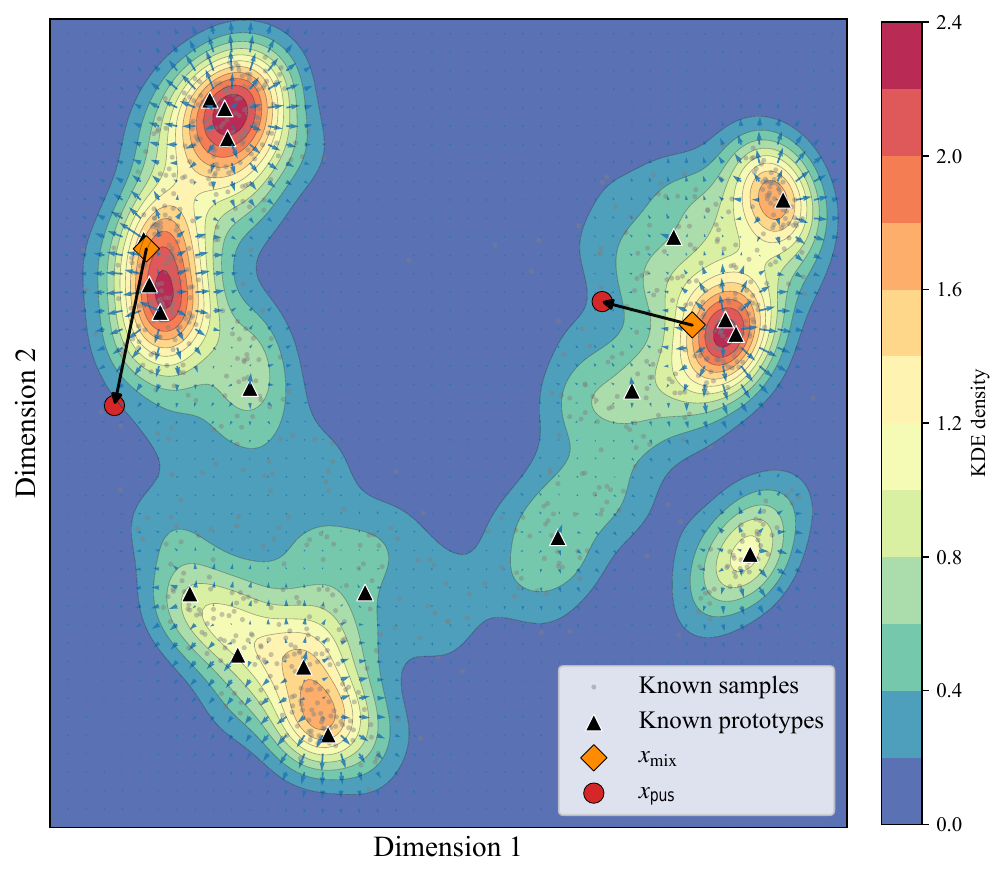}
        \vspace{-0.8mm}
        \small (a) Perturbation direction.
    \end{minipage}\hspace{0.01\linewidth}%
    \begin{minipage}[t]{0.46\linewidth}
        \centering
        \includegraphics[width=\linewidth]{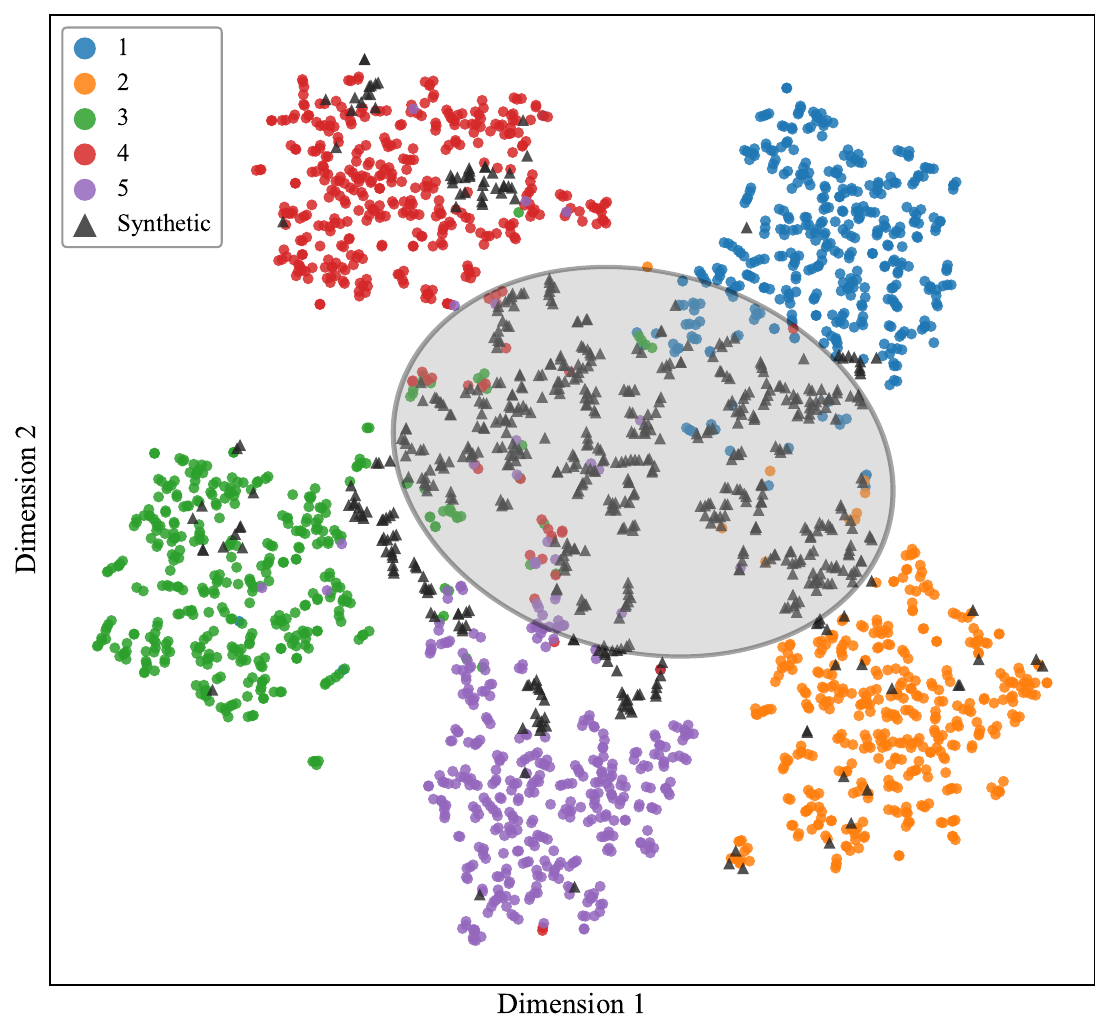}
        \vspace{-0.8mm}
        \small (b) t-SNE on CIFAR-10.
    \end{minipage}
    \vspace{-1mm}
    \caption{\textbf{MKEE generates semantically meaningful pseudo-unknowns.} (a) Visualization of the feature space on the Oxford Pets dataset, illustrating known-class prototypes and the perturbation trajectories from $x_{\text{mix}}$ to $x_{\text{pus}}$.
(b) t-SNE visualization on the CIFAR-10 dataset, where five colored clusters correspond to the five known classes (1–5), and pseudo-unknowns (black triangles) clearly diverge from all known-class regions.}
    \label{fig:ke_tsne}
\end{figure}

\noindent\textbf{MKEE perturbation towards Pseudo-Novel regions.}  
Although \(x_{\text{mix}}\) lies near the class boundaries, it often remains within high-confidence regions of the known distribution.  
To shift these samples to off-support, novel-like areas, we design an objective that jointly promotes prediction uncertainty and discourages proximity to known-class features:
\begin{equation}
\label{eq:kde-global}
\mathcal{J}(x) =
- \sum_{c=1}^{K} p_c \log p_c
- \lambda_{\rho} \cdot \rho(x),
\end{equation}
\noindent
where \(p_c = \frac{\exp(\ell_c(x))}{\sum_k \exp(\ell_k(x))}\) denotes the softmax probability for class \(c\), and  
\(\rho(x) = \frac{1}{N}\sum_{n=1}^{N}\exp\!\left(-\frac{\|f(x)-f(x_n)\|^2}{2\sigma^2}\right)\) estimates the local feature density of known classes via kernel density estimation (KDE)~\cite{chen2017tutorial}.  
Here, \(\lambda_{\rho}\) controls the trade-off between entropy maximization and density minimization.
The first term encourages classifier uncertainty~\cite{ye2025towards}, whereas the second term pushes samples away from high-density regions of known classes. To efficiently explore pseudo novel class directions, we apply a single gradient ascent step on \(\mathcal{J}(x)\):
\begin{equation}
\label{eq:one-step}
\begin{aligned}
x_{\text{pus}} = x_{\text{mix}} + \varepsilon \cdot \left. \frac{\nabla_x \mathcal{J}(x)}{\|\nabla_x \mathcal{J}(x)\|_2} \right|_{x = x_{\text{mix}}}
\end{aligned}
\end{equation}
where \(\varepsilon\) controls the perturbation magnitude.  
As illustrated in Figure~\ref{fig:ke_tsne}(a), this update effectively moves samples from the dense known regions toward sparse, uncertain areas, mimicking the characteristics of unknown categories.

\noindent\textbf{Batch-Wise Density Approximation.}  
Directly computing $\rho(x)$ over the entire training set is computationally expensive (\(\mathcal{O}(N^2)\))~\cite{yang2003improved} and requires manual bandwidth tuning.  
To make this process efficient and adaptive, we estimate the local density of each sample within a mini-batch using a small reference feature set.  
Let \(f(x)\) denote the unit-normalized feature of \(x\), and \(\{f_r^{\text{ref}}\}_{r=1}^{R}\) represent unit features from a reference set (typically drawn from the current batch).  
The local batch-level density is then estimated as:
\begin{equation}
\label{eq:kde-per-sample}
\begin{aligned}
\rho_{\text{batch}}(x)
=\frac{1}{R}\sum_{r=1}^R
\exp\!\left(
-\frac{\| f(x)- f_r^{\text{ref}}\|_2^2}{2\sigma^2}
\right),
\\
\sigma=\sigma_{0}\cdot
\operatorname{median}\!\Big\{\| f_b- f_r^{\text{ref}}\|_2\Big\}_{b\le B,\,r\le R}.
\end{aligned}
\end{equation}

When no external reference set is available, we use the current batch itself, i.e., \(\{f_r^{\text{ref}}\} = \{f_b\}\), for self-referenced estimation.  
Within each batch, multiple mixup anchors are sampled to generate pseudo-novel examples via the one-step update in Eq.~\eqref{eq:one-step}.  
These synthesized pseudo-novel classes are subsequently employed by the dual max-margin objective described in Sec.~\ref{subsec:autoTau}.


\subsection{Dual Max-Margin Loss}
\label{subsec:autoTau}

Our framework performs classification through prototype-based matching in the continuous feature space, as discussed in Sec.~\ref{subsec:baseline}.  
For a given sample \(x\), we compute its cosine similarity to all known-class prototypes and define
\(s_{\max}(x) = \max_k f(x)^\top P_k\)  
as the highest similarity score.  
This score measures how confidently the sample aligns with the known prototype set:  
high values indicate strong association with a known class, whereas low values suggest semantic deviation or potential novelty.
Ideally, samples from known classes should yield high \(s_{\max}(x)\), while pseudo-unknown or truly novel samples should exhibit low similarity scores.  
To explicitly enforce this separation during training, we introduce a \emph{dual max-margin loss} that encourages a clear margin around the decision threshold.

Let \(\mathcal{D}_{\text{known}}\) and \(\mathcal{D}_{\text{pseudo}}\) denote mini-batches of known and pseudo-novel samples, where the latter (\(x_{\text{pus}}\)) are generated using the MKEE perturbation described in Sec.~\ref{subsec:pus}.  
For each batch, we impose margin-based constraints on the similarity scores:
\begin{equation}
\label{eq:Lmm}
\begin{aligned}
\mathcal{L}_{\text{pos}} &= \mathbb{E}_{x\in\mathcal{D}_{\mathrm{known}}}
  \Big[\big[(\tau+m_{\mathrm{pos}})-s_{\max}(x)\big]_+\Big],\\
\mathcal{L}_{\text{neg}} &= \mathbb{E}_{x\in\mathcal{D}_{\mathrm{pseudo}}}
  \Big[\big[s_{\max}(x)-(\tau-m_{\mathrm{neg}})\big]_+\Big],\\
\mathcal{L}_{\text{mm}}  &= \mathcal{L}_{\text{pos}} + \mathcal{L}_{\text{neg}}.
\end{aligned}
\end{equation}
where \([z]_+ = \max(0, z)\).  
This formulation drives known-class samples above a positive margin and pushes pseudo-novel below a negative one, thereby widening the gap around the threshold \(\tau\).  
As a result, the feature space becomes more robust to category shifts during inference.

\noindent\textbf{Adaptive threshold estimation.}
As training progresses, model confidence on known classes typically increases~\cite{ye2026asmil}, rendering a fixed decision boundary suboptimal for distinguishing novel categories.  
To address this, we employ a quantile-based adaptive strategy that dynamically updates the threshold $\tau$ based on the current distribution of prototype-matching scores.  
Specifically, at each iteration we compute the upper quantile of similarity scores from known samples and the lower quantile from pseudo-novel samples:
\begin{equation}
\begin{aligned}
u_{\text{pos}} &= \operatorname{percentile}\big(s_{\max}(x),  q_{\text{pos}}\big), \quad x \in \mathcal{D}_{\text{known}}, \\
u_{\text{neg}} &= \operatorname{percentile}\big(s_{\max}(x),  q_{\text{neg}}\big), \quad x \in \mathcal{D}_{\text{pseudo}}.
\end{aligned}
\end{equation}

The quantiles $q_{\text{pos}}$ and $q_{\text{neg}}$ are hyperparameters controlling the percentile range, e.g., $q_{\text{pos}}=0.8$ and $q_{\text{neg}}=0.2$ use the top 20\% most confident known samples and the bottom 20\% least confident pseudo-novels to guide threshold adjustment.  
A target threshold is set as the midpoint between these quantiles,
$\tau_{\text{target}} = \tfrac{1}{2}(u_{\text{pos}} + u_{\text{neg}})$. 
To ensure that the threshold evolves smoothly with model confidence and remains robust to initialization and distributional shifts, we update $\tau$ using  exponential moving average (EMA)~\cite{morales2024exponential}:
\begin{equation}
\label{eq:autotau}
\tau \leftarrow (1 - \beta)\tau + \beta\tau_{\text{target}}.
\end{equation}

\noindent\textbf{Overall objective.}  Finally, we integrate the margin-based separation loss with the feature learning objectives to form the complete training loss:
\begin{equation}
\label{eq:total_loss}
\mathcal{L}_{\text{total}} = \mathcal{L}_{\text{ce}} + \alpha \mathcal{L}_{\text{sup}} + \gamma_{mm} \mathcal{L}_{\text{mm}} ,
\end{equation}
where $\alpha$ and $\gamma_{mm}$ balance the contributions of the contrastive and margin-based components.  
During inference, the model follows the same prototype-matching strategy as described in Sec.~\ref{subsec:baseline}:  
each test sample is assigned to its nearest prototype, and a new prototype is created whenever $s_{\max}(x) < \tau$.  
The complete training and inference process is summarized in Algorithm~\ref{alg:ltc}.
\begin{algorithm}[h]
\small
\DontPrintSemicolon
\caption{LTC: Training with MKEE and Online Inference}
\label{alg:ltc}
\KwIn{$\mathcal D_S$; encoder $z(\cdot)$; head; hyperparams\{$\alpha,\gamma_{mm},\beta$, $p_{\text{gen}}{=}0.3$ $\text{et al.}$.\} }
\KwOut{$z(\cdot)$, dictionary $\mathrm{Dict}$, threshold $\tau$.}
\KwInit{Prototypes $P_k$ = unit-normalized class means on $\mathcal D_S$; $\mathrm{Dict}=\{P_k\}_{k=1}^K$; $\tau\!\leftarrow\!\tau_{\text{init}}$.}

\textbf{Training:} \For{$epoch=1..E$}{
\For{mini-batch $\mathcal B=\{(x_i,y_i)\}_{i=1}^B$}{
1) $f_i=\frac{z(x_i)}{\|z(x_i)\|_2}$, logits $\ell(x_i)$; compute $\mathcal L_{\text{sup}}$ and $\mathcal L_{\text{ce}}$.\\
2) Draw a batch trigger $h\sim\mathrm{Bernoulli}(p_{\text{gen}})$; \emph{if $h{=}1$}, form $\{x_{\text{mix}}^{(b)}\}_{b=1}^B$ by mixup ($\alpha$); \emph{else} set $\{x_{\text{mix}}^{(b)}\}\!=\!\emptyset$.\\
3) \emph{If $h{=}1$}, obtain $\{x_{\text{pus}}^{(b)}\}_{b=1}^B$ via $x_{\text{pus}}=x_{\text{mix}}+\varepsilon\,\nabla_x\mathcal J(x)/\|\nabla_x\mathcal J(x)\|_2$ by Eq.~\eqref{eq:one-step}; \emph{else} set $\{x_{\text{pus}}^{(b)}\}\!=\!\emptyset$.\\
4) With $s_{\max}(x)=\max_k f(x)^\top P_k$ for $x\in\{x_i\}\cup\{x_{\text{pus}}^{(b)}\}$, compute $\mathcal L_{\text{mm}}$ (Eq.~\eqref{eq:Lmm}); \emph{update $\tau$ (Eq.~\eqref{eq:autotau}) only if $h{=}1$}.\\
5) \textbf{Minimize} the total loss $\mathcal L_{\text{total}}$ in Eq.~\eqref{eq:total_loss}; update $z(\cdot)$, head; update $P_k$ by normalized running mean on labeled $(x_i,y_i)$.
}
}

\textbf{Online inference (OCD stream $\mathcal D_Q$):}
\For{incoming $x$}{
$f(x)$, $s_k=f(x)^\top P_k$, $s_{\max}=\max_k s_k$; 
\\
\If{$s_{\max}<\tau$}{append $P_{K+1}\!\leftarrow\! f(x)$; $K\!\leftarrow\!K+1$}\Else{$\hat y=\arg\max_k s_k$}
}
\end{algorithm}

\section{Experiments}

\begin{table*}[htbp]
\centering
\small
\setlength{\tabcolsep}{5pt}
\renewcommand{\arraystretch}{1.10} 
\caption{\textbf{Comparison with other SOTA methods.} 
“DiffGRE-P” and “DiffGRE-S” represent the integration of DiffGRE with SMILE and PHE, respectively. 
}
\label{tab:sota_comparison}
\resizebox{\textwidth}{!}{
\begin{tabular}{cc
|ccc|ccc|ccc|ccc|ccc|ccc|ccc} 
\toprule
\multirow{2}{*}{ } 
& \multirow{2}{*}[-0.6ex]{\textbf{Method}}
& \multicolumn{3}{c|}{\textbf{CIFAR-10 (\%)}} 
& \multicolumn{3}{c|}{\textbf{CIFAR-100 (\%)}} 
& \multicolumn{3}{c|}{\textbf{ImageNet-100 (\%)}}
& \multicolumn{3}{c|}{\textbf{CUB-200-2011 (\%)}} 
& \multicolumn{3}{c|}{\textbf{Stanford Cars (\%)}} 
& \multicolumn{3}{c|}{\textbf{Oxford Pets (\%)}} 
& \multicolumn{3}{c}{\textbf{Food101 (\%)}} \\
\cmidrule(lr){3-5}\cmidrule(lr){6-8}\cmidrule(lr){9-11}\cmidrule(lr){12-14}\cmidrule(lr){15-17}\cmidrule(lr){18-20}\cmidrule(lr){21-23}
&
& All & Old & New
& All & Old & New
& All & Old & New
& All & Old & New
& All & Old & New
& All & Old & New
& All & Old & New \\
\midrule
\multirow{9}{*}{\rotatebox[origin=c]{90}{\textit{Greedy--Hungarian}}}
& SLC \cite{hartigan1975clustering}
& 65.9 & 96.5 & 50.9
& 46.9 & 62.1 & 16.6
& 34.2 & 86.6 & 7.1
& 31.3 & 48.5 & 22.7
& 24.0 & 45.8 & 13.6
& --   & --   & --
& --   & --   & --
 \\
& MLDG \cite{MLDGli2018learning}
& 71.6 & 97.5 & 58.6
& 58.4 & 69.0 & 37.3
& 33.6 & 74.4 & 13.1
& 34.2 & 57.9 & 22.4
& 28.0 & 49.1 & 17.7
& -- & -- & --
& -- & -- & -- \\
& RankStat \cite{RankStat}
& 56.5 & 81.1 & 44.2
& 36.9 & 45.7 & 19.3
& 33.1 & 74.2 & 12.4
& 27.6 & 46.2 & 18.3
& 18.6 & 36.9 & 9.7
& --   & --   & --
& --   & --   & -- 
 \\
& WTA \cite{WTAjia2021joint}
& 65.4 & 88.0 & 54.1
& 44.1 & 55.5 & 21.2
& 33.1 & 75.8 & 11.7
& 26.5 & 45.0 & 17.3
& 20.0 & 38.8 & 10.6
& --   & --   & --
& --   & --   & -- 
\\
& SMILE \cite{SMILE}
& 78.2 & \textbf{99.3} & 67.6
& 61.3 & 70.7 & 42.5
& 39.9 & 87.1 & 16.2
& 41.1 & 67.6 & 27.8
& 33.4 & 58.4 & 21.3
& 54.1 & 66.1 & 47.8
& 34.4 & 64.0 & 19.4 \\

& PHE-DINO \cite{PHE}
& 83.0 & 98.0 & 75.5
& 64.8 & 78.8 & 36.9
& 53.1 & 83.5 & 38.1
& 46.9 & 76.0 & 32.4
& 46.3 & 78.3 & 30.8
& 63.3 & 91.3 & 48.6
& 50.0 & 89.3 & 30.0 \\

& PHE-CLIP \cite{PHE}
& 79.3 & 97.0 & 70.4
& 66.1 & 80.3 & 37.5
& 52.9 & 87.8 & 35.5
& 44.2 & 70.3 & 31.1
& 46.4 & 78.1 & 31.1
& 64.1 & 86.2 & 52.4
& 47.8 & 88.4 & 27.0 \\
&  \textbf{LTC-DINO(ours)}
& \cellcolor{ltgray}80.5 & \cellcolor{ltgray}98.1 & \cellcolor{ltgray}71.7
& \cellcolor{ltgray}66.8 & \cellcolor{ltgray}81.2 & \cellcolor{ltgray}37.8
& \cellcolor{ltgray}54.0 & \cellcolor{ltgray}\textbf{90.5} & \cellcolor{ltgray}35.7
& \cellcolor{ltgray}51.9 & \cellcolor{ltgray}82.9 & \cellcolor{ltgray}36.3
& \cellcolor{ltgray}42.3 & \cellcolor{ltgray}79.5 & \cellcolor{ltgray}24.4
& \cellcolor{ltgray}68.6 & \cellcolor{ltgray}92.5 & \cellcolor{ltgray}56.0
& \cellcolor{ltgray}43.0 & \cellcolor{ltgray}82.2 & \cellcolor{ltgray}23.0 \\
& \textbf{LTC-CLIP(ours)}
& \cellcolor{ltgray}\textbf{88.6} & \cellcolor{ltgray}98.1 & \cellcolor{ltgray}\textbf{83.8}
& \cellcolor{ltgray}\textbf{70.7} & \cellcolor{ltgray}\textbf{81.5} & \cellcolor{ltgray}\textbf{49.3}
& \cellcolor{ltgray}\textbf{55.6} & \cellcolor{ltgray}87.7 & \cellcolor{ltgray}\textbf{39.5}
& \cellcolor{ltgray}\textbf{57.8} & \cellcolor{ltgray}\textbf{83.9} & \cellcolor{ltgray}\textbf{44.8}
& \cellcolor{ltgray}\textbf{56.6} & \cellcolor{ltgray}\textbf{90.3} & \cellcolor{ltgray}\textbf{40.4}
& \cellcolor{ltgray}\textbf{73.0} & \cellcolor{ltgray}\textbf{92.6} & \cellcolor{ltgray}\textbf{62.7}
& \cellcolor{ltgray}\textbf{54.7} & \cellcolor{ltgray}\textbf{90.7} & \cellcolor{ltgray}\textbf{36.4} \\
\midrule
\multirow{11}{*}{\rotatebox[origin=c]{90}{\textit{Strict--Hungarian}}}
& SLC \cite{hartigan1975clustering}
& 41.5 & \textbf{58.3} & 33.3
& 44.4 & 59.0 & 15.1
& 32.9 & \textbf{86.6} & 5.2
& 28.6 & 44.0 & 20.9
& 14.0 & 23.0 & 9.7
& 35.5 & 41.3 & 33.1
& 20.9 & 48.6 & 6.8
 \\
& MLDG \cite{MLDGli2018learning} 
& 44.1 & 38.5 & 47.0
& 50.6 & 61.0 & 29.8
& 30.6 & 72.3 & 9.7
& 29.5 & 48.4 & 20.1
& 24.0 & 41.6 & 15.4
& --   & --   & --
& --   & --   & -- \\
& RankStat \cite{RankStat}
& 42.1 & 49.3 & 38.6
& 35.0 & 44.0 & 17.0
& 31.1 & 73.3 & 9.8
& 21.2 & 26.9 & 18.4
& 14.8 & 19.9 & 12.3
& 33.2 & 42.3 & 28.4
& 22.3 & 50.7 & 7.8
\\
& WTA \cite{WTAjia2021joint}
& 43.1 & 34.5 & 47.4
& 40.8 & 52.9 & 16.7
& 30.8 & 72.9 & 9.7
& 21.9 & 26.9 & 19.4
& 17.1 & 24.4 & 13.6
& 35.2 & 46.3 & 29.3
& 18.2 & 40.5 & 6.1 
\\
& SMILE \cite{SMILE}
& 49.9 & 39.9 & 54.9
& 51.6 & 61.6 & 31.7
& 33.8 & 74.2 & 13.5
& 32.2 & 50.9 & 22.9
& 26.2 & 46.6 & 16.3
& 42.9 & 38.7 & 45.1
& 24.2 & 54.3 & 8.8 \\
& PHE-DINO \cite{PHE}
& 53.1 & 19.3 & 70.0
& 56.0 & 70.1 & 27.8
& 39.2 & 49.3 & 34.1
& 36.4 & \textbf{55.8} & 27.0
& 31.3 & 61.9 & 16.8
& 48.3 & 53.8 & 45.4
& 29.1 & 64.7 & 11.1 \\
& PHE-CLIP \cite{PHE}
& 52.4 & 18.3 & 69.5
& 56.8 & 71.9 & 26.5
& 39.2 & 60.7 & 28.4
& 35.1 & 54.5 & 25.4
& 36.2 & 54.2 & 27.4
& 52.0 & 52.3 & 51.9
& 33.5 & 58.6 & 20.6 \\
& DiffGRE-S~\cite{DiffGRE} & -- & -- & -- & -- & -- & -- & -- & -- & -- & 35.4 & 58.2 & 23.8 & 30.5 & 59.3 & 16.5 & 42.4 & 42.1 & 42.5 & -- & -- & -- \\
& DiffGRE-P~\cite{DiffGRE} & -- & -- & -- & -- & -- & -- & -- & -- & -- & 37.9 & 57.0 & 28.3 & 32.1 & 63.3 & 16.9 & 48.6 & 52.6 & 46.6 & -- & -- & -- \\
& \textbf{LTC-DINO(ours)}
& \cellcolor{ltgray}53.9 & \cellcolor{ltgray}19.5 & \cellcolor{ltgray}71.1
& \cellcolor{ltgray}57.7 & \cellcolor{ltgray}\textbf{75.1} & \cellcolor{ltgray}22.8
& \cellcolor{ltgray}41.0 & \cellcolor{ltgray}61.5 & \cellcolor{ltgray}30.7
& \cellcolor{ltgray}35.9 & \cellcolor{ltgray}52.4 & \cellcolor{ltgray}27.6
& \cellcolor{ltgray}32.6 & \cellcolor{ltgray}59.8 & \cellcolor{ltgray}19.4
& \cellcolor{ltgray}52.3 & \cellcolor{ltgray}52.2 & \cellcolor{ltgray}52.4
& \cellcolor{ltgray}29.7 & \cellcolor{ltgray}66.6 & \cellcolor{ltgray}10.9 \\
& \textbf{LTC-CLIP(ours)}
& \cellcolor{ltgray}\textbf{54.6} & \cellcolor{ltgray}19.3 & \cellcolor{ltgray}\textbf{72.3}
& \cellcolor{ltgray}\textbf{60.0} & \cellcolor{ltgray}73.5 & \cellcolor{ltgray}\textbf{32.8}
& \cellcolor{ltgray}\textbf{45.9} & \cellcolor{ltgray}67.9 & \cellcolor{ltgray}\textbf{35.0}
& \cellcolor{ltgray}\textbf{42.5} & \cellcolor{ltgray}51.7 & \cellcolor{ltgray}\textbf{37.8}
& \cellcolor{ltgray}\textbf{49.3} & \cellcolor{ltgray}\textbf{74.0} & \cellcolor{ltgray}\textbf{37.4}
& \cellcolor{ltgray}\textbf{58.9} & \cellcolor{ltgray}\textbf{59.5} & \cellcolor{ltgray}\textbf{58.5}
& \cellcolor{ltgray}\textbf{37.6} & \cellcolor{ltgray}\textbf{72.3} & \cellcolor{ltgray}\textbf{36.7} \\
\bottomrule
\end{tabular}
}
\end{table*}

\subsection{Experimental Setup}
\noindent\textbf{Datasets.}  
We evaluate on four fine-grained benchmarks (CUB~\cite{CUB}, Stanford Cars~\cite{scars}, Oxford Pets~\cite{pets}, Food-101~\cite{food}) and three coarse-grained datasets (CIFAR-10/100~\cite{cifar}, ImageNet-100~\cite{2015imagenet}). Following~\cite{SMILE,PHE}, each dataset is split into seen and unseen categories, with 50\% of the seen samples used for training and the rest for evaluation. Full split statistics are provided in the \textit{Supp.}

\noindent\textbf{Evaluation and implementation details.}  Following the standard evaluation protocol in \cite{SMILE}, we evaluate performance with clustering accuracy (ACC), including both the Strict--Hungarian and Greedy--Hungarian variants. We adopt both CLIP-pretrained~\cite{CLIP} and DINO-pretrained~\cite{caron2021emerging} ViT-B/16 as the backbone (more results are provided in the \textit{Supp}). In line with PHE~\cite{PHE} and SMILE~\cite{SMILE}, we fine-tune only the last transformer block of ViT-B/16 while freezing all remaining parameters. The loss weights $\alpha$ and $\gamma_{mm}$ are fixed to 0.3 and 0.05 across all datasets. As detailed in Sec.~\ref{subsec:autoTau}, the threshold $\tau$ is adaptively updated from an initialization $\tau_{\text{init}}$; we set $\tau_{\text{init}}{=}0.7$ for CIFAR-10, CIFAR-100, ImageNet-100, and Oxford Pets, and $\tau_{\text{init}}{=}0.6$ for CUB, Stanford Cars, and Food-101. We use $m_{\text{pos}}{=}m_{\text{neg}}{=}0.05, \beta=0.001$ for all datasets. Additional implementation details are included in the \textit{Supp.}

\noindent\textbf{Compared Methods.}  
We compare with recent OCD methods including SMILE~\cite{SMILE}, PHE~\cite{PHE}, and DiffGRE~\cite{DiffGRE}, as well as four baselines: SLC~\cite{hartigan1975clustering}, RankStat~\cite{RankStat}, WTA~\cite{WTAjia2021joint}, and MLDG~\cite{MLDGli2018learning}. Details and configurations follow prior work and are provided in the \textit{Supplement}.

\subsection{Comparison with State of the Art}

We evaluate our method against state-of-the-art approaches across seven datasets, with results summarized in Table~\ref{tab:sota_comparison}. On all-class accuracy, LTC outperforms the strongest baseline by an average of 5.54\%, with gains ranging from 1.5\% (CIFAR-10) to 13.1\% (SCars). Improvements are especially notable on fine-grained datasets (7.84\% average across four), highlighting LTC’s strength in modeling subtle semantic differences. In contrast, coarse-grained datasets like CIFAR offer limited headroom, resulting in smaller gains. While overall trends under both evaluation protocols align, exceptions exist. For instance, on CIFAR-10, LTC, PHE, and SMILE perform well under the Greedy protocol but fall below 50\% old-class accuracy under Strict evaluation. As noted by SMILE~\cite{SMILE}, this is due to CIFAR-10’s low class diversity, which biases assignment toward dominant novel classes. Table~\ref{tab:pred_acc_count} further shows that LTC yields lower category count estimation error than PHE and SMILE, helping mitigate “category explosion.”

\begin{table}[tbp]
\small
\centering
\setlength{\tabcolsep}{2pt}
\caption{\textbf{Comparison of estimated category numbers and accuracies on the CUB-200-2011 and Stanford Cars datasets.} }
\label{tab:pred_acc_count}
\resizebox{0.48\textwidth}{!}{
\begin{tabular}{l|c|ccc|c|ccc}
\toprule
\multirow{2}{*}[-0.6ex]{Method}
& \multicolumn{4}{c|}{CUB ($C{=}200$)} 
& \multicolumn{4}{c}{SCars ($C{=}196$)} \\
\cmidrule(lr){2-5} \cmidrule(lr){6-9}
& \#Cls & All & Old & New & \#Cls & All & Old & New \\
\midrule
SMILE-16bit~\cite{SMILE}    
& 924 & 31.9 & 52.7 & 21.5 & 896 & 27.5 & 52.5 & 15.4 \\
SMILE-32bit~\cite{SMILE}    
& 2146 & 27.3 & 52.0 & 15.0 & 2953 & 21.9 & 46.8 & 9.9 \\
PHE-16bit~\cite{PHE}        
& 318 & 37.6 & 57.4 & 27.6 & 709 & 31.8 & 65.4 & 15.6 \\
PHE-32bit~\cite{PHE}        
& 474 & 38.5 & 59.9 & 27.8 & 762 & 31.5 & 64.0 & 15.8 \\
\rowcolor{gray!10}
\textbf{LTC (ours)}    & 
210
& 57.8 & 83.9 & 44.8 & 389 & 56.6 & 90.3 & 40.4 \\
\bottomrule
\end{tabular}
}
\end{table}

A key observation lies in the contrast across backbones: prior methods like PHE and SMILE show limited gains when switching from DINO- to CLIP-pretrained ViT, while our LTC achieves substantial improvement. This stems from two factors: (i) hash-based pipelines compress features into binary codes, discarding semantic and fine-grained cues; (ii) training only on known classes restricts the use of CLIP’s semantic priors. In contrast, LTC adopts a fully feature-based design and introduces generative supervision via pseudo-unknowns, enhancing its ability to model high-level semantics and category boundaries. This aligns with~\cite{liu2025data}, which shows that CLIP, under equal conditions, yields richer semantic representations than DINO. By unlocking this potential, LTC excels in fine-grained discovery. We therefore recommend CLIP-pretrained ViT as a standard backbone for OCD research, enabling fairer comparisons and pushing the performance frontier.

\subsection{Ablation study and analysis}
\begin{table}[tbp]
  \setlength{\tabcolsep}{6pt}
  \footnotesize
  \renewcommand{\arraystretch}{0.9}
  \centering
  \caption{\textbf{Ablation study of key components in LTC.}  Results on Oxford Pets and Stanford Cars under the Greedy–Hungarian protocol. $\mathcal{L}_{\text{ce}}$, $\mathcal{L}_{\text{sup}}$, and $\mathcal{L}_{\text{mm}}$ correspond to the cross-entropy, supervised contrastive, and dual max-margin losses, respectively, while “$w/o$ MKEE” removes the pseudo-unknown generation module.}
  \label{tab:ab_com}
  \resizebox{0.48\textwidth}{!}{
  \begin{tabular}{l|ccc|ccc}
    \toprule
    \multirow{2}{*}[-0.8ex]{Method} & \multicolumn{3}{c|}{Pets (\%)} & \multicolumn{3}{c}{SCars (\%)} \\
    \cmidrule(lr){2-4} \cmidrule(lr){5-7}
    & All & Old & New & All & Old & New \\
    \midrule
    $w/o$ $\mathcal{L}_{\text{ce}}$              
    & 71.2 & 94.0 & 59.2 & 55.4 & 90.2 & 38.6 \\
    $w/o$ $\mathcal{L}_{\text{sup}}$             
    & 67.6 & 89.5 & 56.2 & 51.9 & 84.4 & 36.2 \\
    $w/o$ $\mathcal{L}_{\text{mm}}$     
    & 68.5 & 94.4 & 54.9 & 55.8 & 90.1 & 39.3 \\
    $w/o$ MKEE                       
    & 68.8 & 95.4 & 54.8 & 55.2 & 87.7 & 39.4 \\
    \rowcolor{gray!10}
    \textbf{LTC (ours)}                  
    & 73.0 & 92.6 & 62.7 & 56.6 & 90.3 & 40.4 \\
    \bottomrule
  \end{tabular}
  }
\end{table}
\noindent\textbf{Components Ablation. } We perform ablation on the Oxford-IIIT Pet and Stanford Cars datasets, as summarized in Table~\ref{tab:ab_com}. To assess the role of $\mathcal{L}_{ce}$ and $\mathcal{L}_{sup}$ in representation learning, we remove each loss individually. Results show that both are essential for learning discriminative features. We then examine the effect of the max-margin loss $\mathcal{L}_{mm}$. As defined in Eq.~\ref{eq:Lmm}, $\mathcal{L}_{mm}$ consists of two parts: one encouraging high scores for known-class samples, and the other suppressing scores for synthesized pseudo-unknowns. We test two ablations: removing $\mathcal{L}_{mm}$ entirely, and keeping only the known-class term.  Results show that incorporating pseudo-unknowns during training is critical, especially for discovering new categories. If $\mathcal{L}_{mm}$ only promotes high confidence on known classes, the model may overfit to seen categories and suffer significantly in its ability to detect novel ones.


\begin{table}[!bp]
\centering
\small
\setlength{\tabcolsep}{9pt}
\caption{\textbf{Comparison of different pseudo-unknown generation methods.}
Results are reported on the Pets dataset under the Greedy–Hungarian protocol.
Synthesis time denotes the cost of generating 128 samples.
MKEE achieves the highest all and novel-class accuracy with negligible computational overhead ($<1$s).}

\label{tab:synthesis_comparison}
\resizebox{0.48\textwidth}{!}{
\begin{tabular}{l|ccc|c}
\toprule
Method & All & Old & New & Time $\downarrow$ \\
\midrule
Mixup\cite{zhang2017mixup} & 71.5 & 96.2 & 58.5 & $<$1s \\
CutMix~\cite{yun2019cutmix} & 69.4& 94.2 & 56.3 & $<$1s 
\\
DiffGRE~\cite{DiffGRE} & 69.9 & 94.6 & 57.0 & 283.96s \\
\midrule
MKEE w/o Entropy & 72.3 & 95.8 & 59.9 & $<$1s \\
MKEE w/o KDE     & 72.2 & 94.2 & 60.6 & $<$1s \\
\rowcolor{gray!10}
\textbf{MKEE (ours)} & 73.0 & 92.6 & 62.7 & $<$1s 
\\
\bottomrule
\end{tabular}
}
\end{table}

\begin{figure}[tbp]
  \centering
  \begin{subfigure}[t]{0.48\linewidth}
    \centering
    \includegraphics[width=\linewidth]{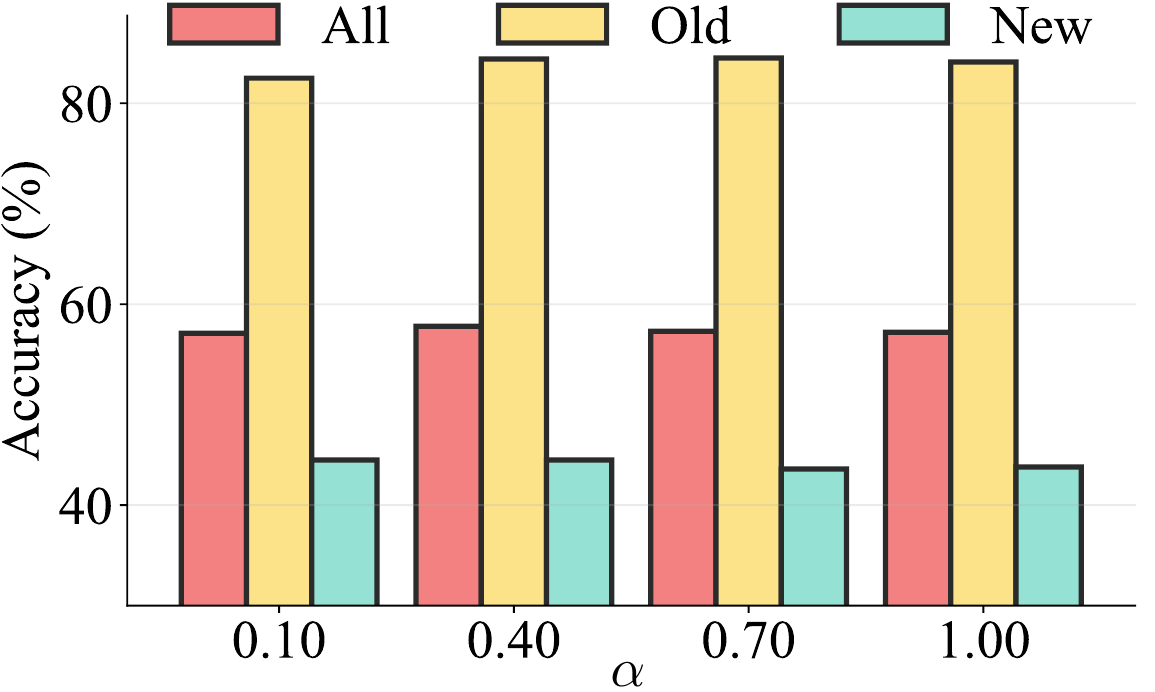}
    \caption{CUB @ $\gamma_{mm}=0.05$}
    \label{fig:cub-sens-alpha}
  \end{subfigure}\hfill
  \begin{subfigure}[t]{0.48\linewidth}
    \centering
    \includegraphics[width=\linewidth]{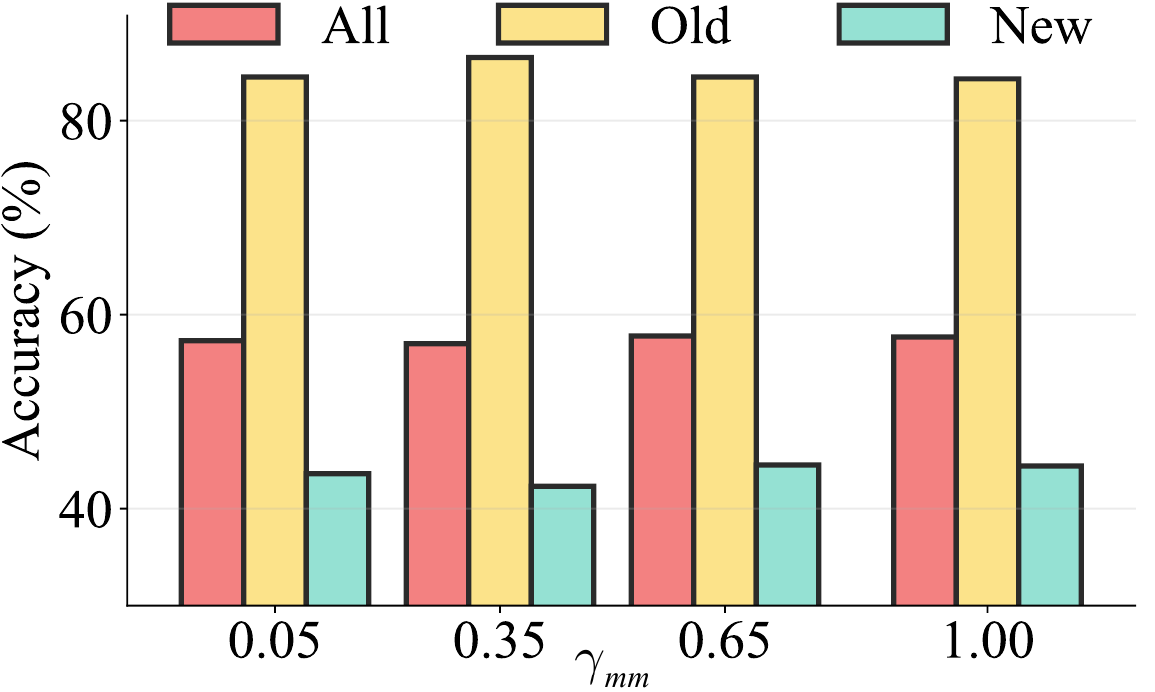}
    \caption{CUB @ $\alpha=0.30$}
    \label{fig:cub-sens-gamma}
  \end{subfigure}

  \vspace{0.6em}

  \begin{subfigure}[t]{0.48\linewidth}
    \centering
    \includegraphics[width=\linewidth]{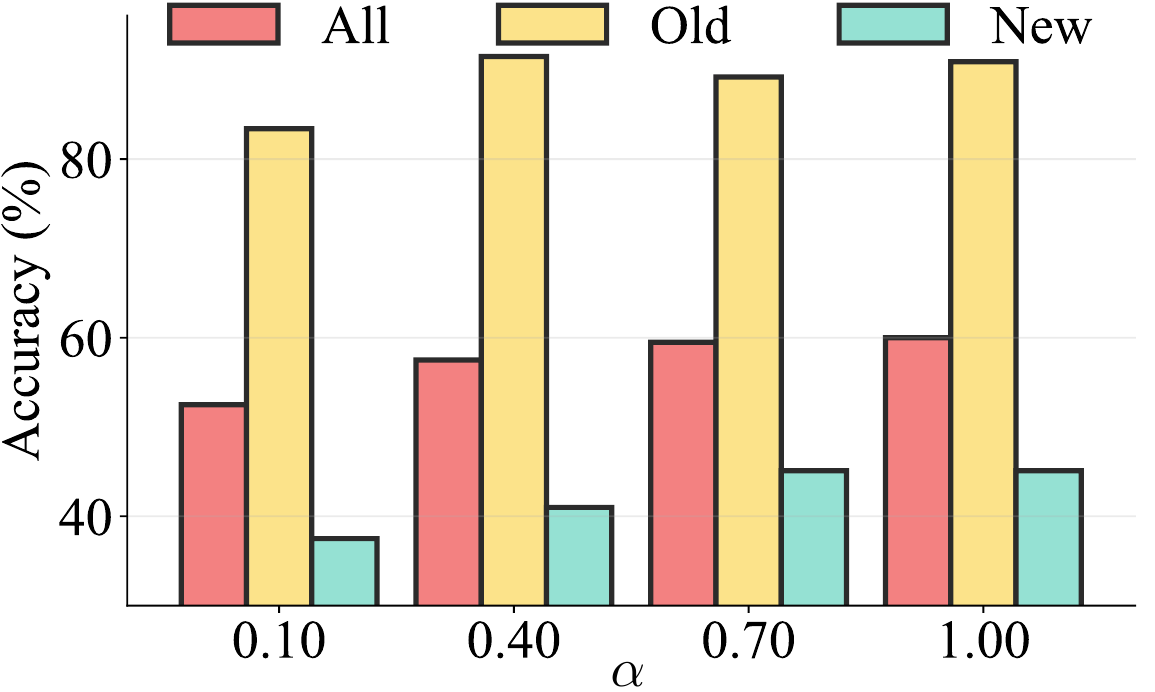}
    \caption{SCars @ $\gamma_{mm}=0.05$}
    \label{fig:scars-sens-alpha}
  \end{subfigure}\hfill
  \begin{subfigure}[t]{0.48\linewidth}
    \centering
    \includegraphics[width=\linewidth]{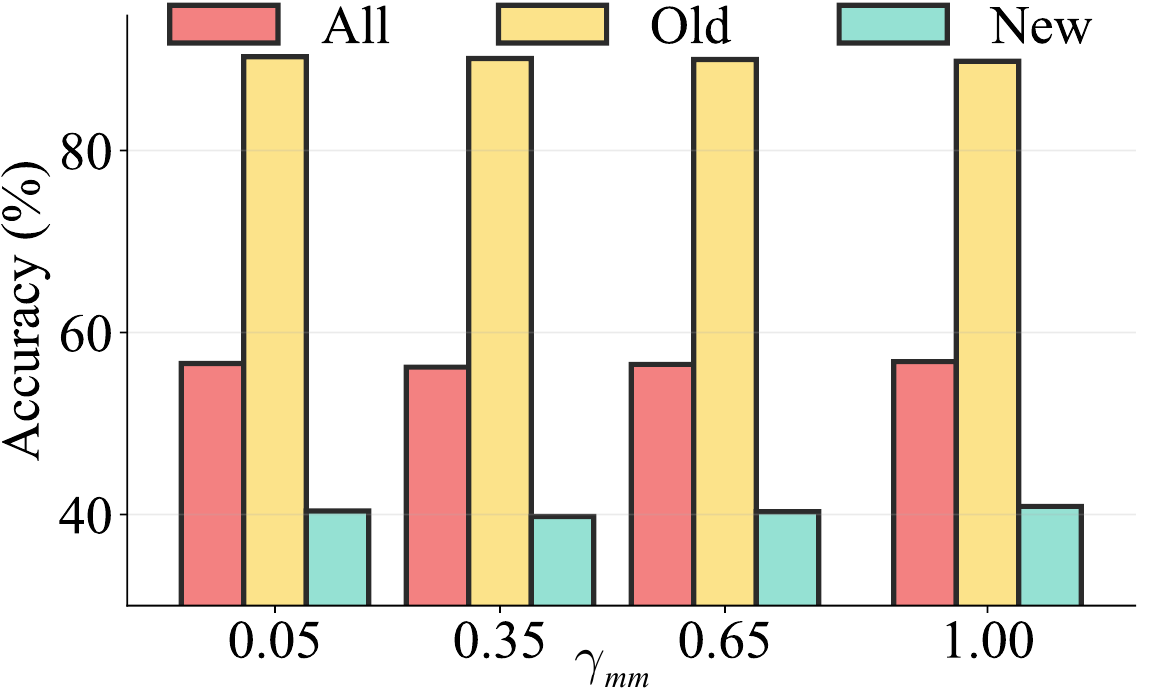}
    \caption{SCars @ $\alpha=0.30$}
    \label{fig:scars-sens-gamma}
  \end{subfigure}

    \caption{\textbf{Hyperparameter sensitivity on the CUB and SCars datasets.} $\alpha$ and $\gamma_{mm}$ control the relative contributions of the contrastive and margin-based terms in the total loss $\mathcal{L}_{\text{total}}$.}   
  \label{fig:grouped-sens-2x2}
\end{figure}

\noindent\textbf{Comparison of Different Pseudo-unknown Generation Methods.} Table~\ref{tab:synthesis_comparison} compares our pseudo-unknown synthesis method with Mixup~\cite{zhang2017mixup}, CutMix~\cite{yun2019cutmix}, and DiffGRE~\cite{DiffGRE}. Mixup and CutMix are simple interpolation methods not designed for novel category discovery, often producing overly smooth or inconsistent samples. DiffGRE uses a diffusion model that generates synthetic samples offline, taking about 284 seconds to generate 128 images. In contrast, our MKEE generates pseudo-unknowns online, producing 30-40 samples per iteration, adapting to the model's evolving understanding. This results in sub-second synthesis time, much more efficient than DiffGRE, which requires generating 1.4K images in advance, taking 40 minutes to an hour. While a direct time comparison is not fully meaningful due to different methods, our approach achieves comparable accuracy with significantly lower computational cost, proving effective and efficient for category discovery.

\begin{figure}[tbp]
  \centering
  \begin{subfigure}[t]{0.48\linewidth}
    \centering
    \includegraphics[width=\linewidth]{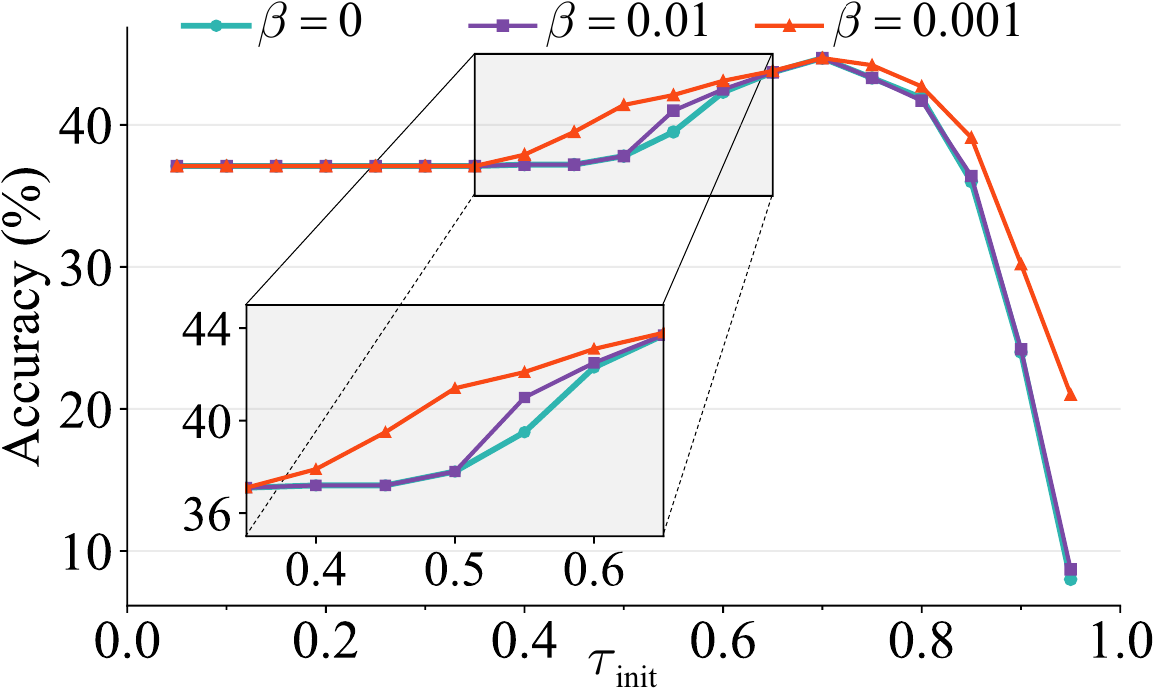}
    \caption{CUB}
    \label{fig:cub-sens-overview}
  \end{subfigure}\hfill
  \begin{subfigure}[t]{0.48\linewidth}
    \centering
    \includegraphics[width=\linewidth]{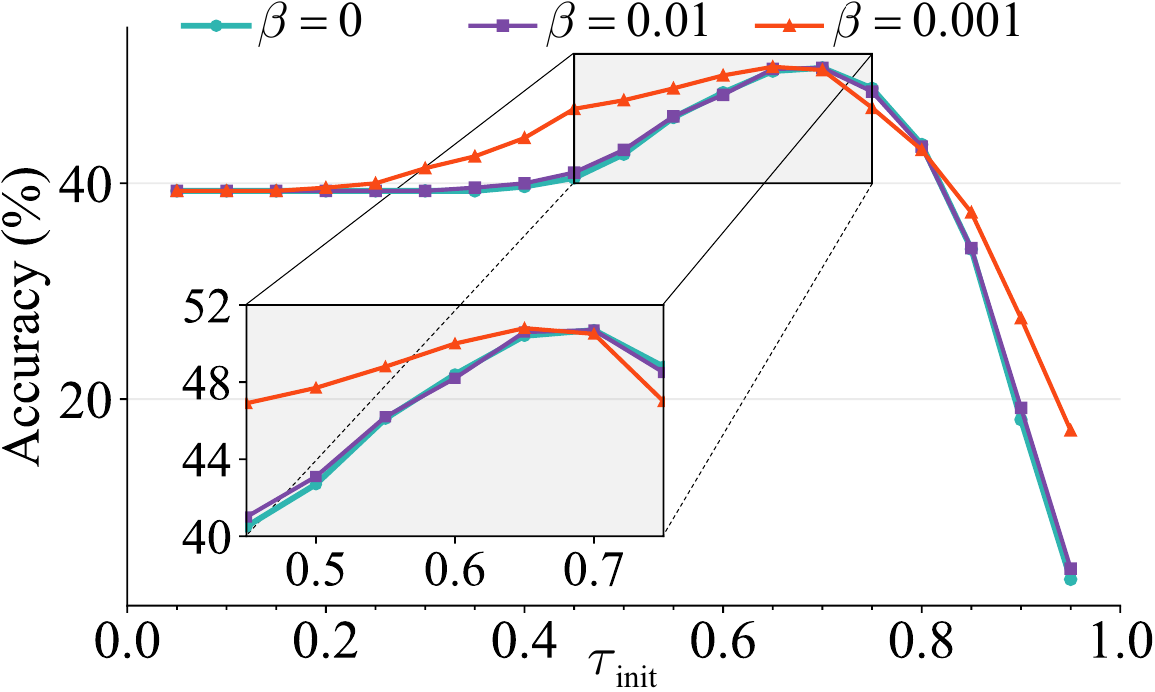}
    \caption{SCars}
    \label{fig:scars-sens-overview}
  \end{subfigure}
\caption{\textbf{Adaptive thresholding improves stability under different initializations.} We study the effect of varying the initial similarity threshold $\tau_{\text{init}}$ and the update rate $\beta$ on overall accuracy.
Results on CUB and SCars show that adaptive updates ($\beta>0$) maintain stable performance across a wide range of $\tau_{\text{init}}$ values, while fixed thresholds ($\beta=0$) are more sensitive to initialization.}
  \label{fig:tau-sens-1x2}
\end{figure}

\noindent\textbf{Hyperparameter Sensitivity. } The influence of the loss weights $\alpha$ and $\gamma_{\text{mm}}$ in Eq.~\ref{eq:total_loss} is shown in Figure~\ref{fig:grouped-sens-2x2}(a)–(d). We fix one coefficient 
while varying the other across a reasonable range: in (a), $\gamma_{\text{mm}}$ is fixed at 0.05; in (b), $\alpha$ is fixed at 0.3. Panels (a)–(d) are evaluated under the Greedy Hungarian protocol, while the $\tau$ curves (Figure~\ref{fig:tau-sens-1x2}) use the strict Hungarian protocol. Overall, the model is more sensitive to changes in $\alpha$, yet the accuracy remains relatively stable across a wide range of values. This robustness indicates that our method does not rely on careful hyperparameter tuning to perform well. We also examine the key parameters in the adaptive threshold update Eq.~\ref{eq:autotau}. Specifically, we vary the growth rate $\beta \in \{0, 0.001, 0.01\}$, and sweep the initial threshold $\tau_{\text{init}}$ from 0 to 1 in steps of 0.05. As shown in Figure~\ref{fig:tau-sens-1x2}(a), adaptive thresholding yields the expected benefits: compared to the static case ($\beta=0$), strategies with $\beta>0$ consistently achieve better performance across a broader range of $\tau_{\text{init}}$, reducing sensitivity to initialization and enabling strong results with less tuning effort. The same trends are observed on Stanford Cars (Figure~\ref{fig:tau-sens-1x2}(b)).

\noindent\textbf{Visualization of Pseudo Unknown Samples.} In Figure~\ref{fig:ke_tsne} (a), we visualize how entropy maximization and kernel-density minimization steer the perturbation direction. The designed perturbation moves the generated samples from the initial mixup state toward more uncertain, out-of-support regions. To further characterize the pseudo-unknowns, Figure~\ref{fig:ke_tsne} (b) shows a t-SNE plot: the pseudo samples (black) are distinct from every known class and occupy portions of the unknown space, enabling the model to better learn what is novel versus known.

\section{Conclusion}
We introduced LTC, a hash-free, creation-driven framework for On-the-Fly Category Discovery. Instead of training only on known classes and hoping for unknown category generalization, LTC exposes the model to pseudo-unknowns during training via MKEE—a one-step entropy–kernel ascent from mixup anchors that uses only the current model and batch features. These created samples are integrated with a dual max-margin objective around an adaptively estimated threshold, yielding a single dynamic prototype dictionary shared by training and inference. Across 7 benchmarks, LTC improves all-class accuracy and unknown discovery.
\section{Acknowledgments}

The paper is supported in part by Beijing Smart Agriculture Innovation Consortium Project (BAIC10-2025). The authors gratefully acknowledge National Innovation Center for Digital Fishery - China Agricultural University, State Key Laboratory of Efficient Utilization of Agricultural Water Resources- China Agricultural University, Key Laboratory of Agricultural Informatization Standardization - MARA, P. R. China, Key Laboratory of Smart Farming Technologies for Aquatic Animals and Livestock - MARA, P. R. China, National Innovation Center for Digital Agricultural Products Circulation - MARA, P. R. China.


{
    \small
    \bibliographystyle{ieeenat_fullname}
    \bibliography{main}
}

\clearpage
\setcounter{page}{1}
\maketitlesupplementary
\appendix

\section*{Appendix Overview}
In this appendix, we provide additional details and experimental results for the main paper. Section~\ref{Backbone} presents a comparative study of CLIP and DINO as backbones for the On-the-Fly Category Discovery (OCD) task. Section~\ref{ARU} discusses the effectiveness of the AutoTau mechanism and other results under varying hyperparameters. Section~\ref{MKEEEEE} offers a theoretical analysis of the Minimizing Kernel Energy \& Maximizing Entropy (MKEE) approach and its impact on known-class accuracy. Finally, Section~\ref{limit} addresses the limitations of our proposed method and suggests potential future improvements.

\section{Backbone Control Study}
\label{Backbone}

To further validate that CLIP serves as a more effective backbone than DINO for On-the-Fly Category Discovery (OCD), we perform additional experiments under two algorithmic variants: the standard Greedy-Hungarian (our main evaluation protocol), and a stricter variant Strict-Hungarian, where pseudo-unknown samples are only used to model novel categories without influencing predictions for known classes. Results are reported in Table~\ref{tab:variants_comparison}.

Across both settings, switching from DINO to CLIP consistently improves performance for SMILE, PHE, and our proposed LTC—especially on the four fine-grained datasets. This trend aligns with findings in Liu et al.~\cite{liu2025data}, which compares CLIP and DINO using identical architectures (ViT-B/16) and training data (ImageNet-style corpus of 100M images). Their study shows that CLIP yields stronger results in fine-grained classification due to its vision-language pretraining, which helps the model capture high-level semantic distinctions between visually similar yet categorically different classes. This capability is particularly well-suited to OCD, where recognizing subtle differences among previously unseen categories is essential.

We therefore suggest building OCD-related tasks upon a CLIP-pretrained backbone, in order to fully leverage its semantic representation capabilities and achieve stronger performance. Future work should also explore model designs that are better aligned with CLIP-pretrained ViT-B/16, aiming to further exploit the rich high-level semantics it encodes and enhance novel category recognition in open-world scenarios.
\begin{table*}[htbp] 
\centering 
\small 
\setlength{\tabcolsep}{5pt} 
\renewcommand{\arraystretch}{1.10} 
\caption{Comparison of variants.} \label{tab:variants_comparison}
\resizebox{\textwidth}{!}{
\begin{tabular}{M{12mm} M{22mm} M{18mm}
|ccc|ccc|ccc|ccc|ccc|ccc|ccc}
\toprule
\multirow{2}{*}{ }
& \multirow{2}{*}[-0.6ex]{\textbf{Method}}
& \multirow{2}{*}[-0.6ex]{\textbf{Backbone}}
& \multicolumn{3}{c|}{\textbf{CIFAR10 (\%)}} 
& \multicolumn{3}{c|}{\textbf{CIFAR100 (\%)}} 
& \multicolumn{3}{c|}{\textbf{ImageNet-100 (\%)}} 
& \multicolumn{3}{c|}{\textbf{CUB-200-2011 (\%)}} 
& \multicolumn{3}{c|}{\textbf{Stanford Cars (\%)}} 
& \multicolumn{3}{c|}{\textbf{Oxford Pets (\%)}} 
& \multicolumn{3}{c}{\textbf{Food101 (\%)}} \\
\cmidrule(lr){4-6}\cmidrule(lr){7-9}\cmidrule(lr){10-12}\cmidrule(lr){13-15}\cmidrule(lr){16-18}\cmidrule(lr){19-21}\cmidrule(lr){22-24}
&
& 
& All & Old & New
& All & Old & New
& All & Old & New
& All & Old & New
& All & Old & New
& All & Old & New
& All & Old & New \\
\midrule
\multirow{6}{*}{\rotatebox[origin=c]{90}{\textit{Greedy--Hungarian}}}
& SMILE & DINO
& 78.2 & 99.3 & 67.6
& 61.3 & 70.7 & 42.5
& 39.9 & 87.1 & 16.2
& 41.1 & 67.6 & 27.8
& 33.4 & 58.4 & 21.3
& 54.1 & 66.1 & 47.8
& 34.4 & 64.0 & 19.4 \\
& SMILE & CLIP
& 82.4 & 97.4 & 74.9
& 56.4 & 64.6 & 40.0
& 47.5 & 71.0 & 35.7
& 43.7 & 69.7 & 30.8
& 36.7 & 57.2 & 26.8
& 58.2 & 77.5 & 48.1
& 40.5 & 70.4 & 25.2 \\
& PHE & DINO
& 83.0 & 98.0 & 75.5
& 64.8 & 78.8 & 36.9
& 53.1 & 83.5 & 38.1
& 46.9 & 76.0 & 32.4
& 46.3 & 78.3 & 30.8
& 63.3 & 91.3 & 48.6
& 50.0 & 89.3 & 30.0 \\
& PHE & CLIP
& 79.3 & 97.0 & 70.4
& 66.1 & 80.3 & 37.5
& 52.9 & 87.8 & 35.5
& 44.2 & 70.3 & 31.1
& 46.4 & 78.1 & 31.1
& 64.1 & 86.2 & 52.4
& 47.8 & 88.4 & 27.0 \\
& LTC & DINO
& 80.5 & 98.1 & 71.7
& 66.8 & 81.2 & 37.8
& 54.0 & 90.5 & 35.7
& 51.9 & 82.9 & 36.3
& 42.3 & 79.5 & 24.4
& 68.6 & 92.5 & 56.0
& 43.0 & 82.2 & 23.0 \\
& LTC & CLIP
& 88.6 & 98.1 & 83.8
& 70.7 & 81.5 & 49.3
& 55.6 & 87.7 & 39.5
& 57.8 & 83.9 & 44.8
& 56.6 & 90.3 & 40.4
& 73.0 & 92.6 & 62.7
& 54.7 & 90.7 & 36.4 \\
\midrule
\multirow{6}{*}{\rotatebox[origin=c]{90}{\textit{Strict--Hungarian}}}
& SMILE & DINO
& 49.9 & 39.9 & 54.9
& 51.6 & 61.6 & 31.7
& 33.8 & 74.2 & 13.5
& 32.2 & 50.9 & 22.9
& 26.2 & 46.6 & 16.3
& 42.9 & 38.7 & 45.1
& 24.2 & 54.3 & 8.8 \\
& SMILE & CLIP
& 51.9 & 19.7 & 68.0
& 46.7 & 55.3 & 29.5
& 35.7 & 41.4 & 32.8
& 34.7 & 55.2 & 24.5
& 32.4 & 46.2 & 25.7
& 40.3 & 37.4 & 41.8
& 33.3 & 56.3 & 21.5 \\
& PHE & DINO
& 53.1 & 19.3 & 70.0
& 56.0 & 70.1 & 27.8
& 39.2 & 49.3 & 34.1
& 36.4 & 55.8 & 27.0
& 31.3 & 61.9 & 16.8
& 48.3 & 53.8 & 45.4
& 29.1 & 64.7 & 11.1 \\
& PHE & CLIP
& 52.4 & 18.3 & 69.5
& 56.8 & 71.9 & 26.5
& 39.2 & 60.7 & 28.4
& 35.1 & 54.5 & 25.4
& 36.2 & 54.2 & 27.4
& 52.0 & 52.3 & 51.9
& 33.5 & 58.6 & 20.6 \\
& LTC & DINO
& 53.9 & 19.5 & 71.1
& 57.7 & 75.1 & 22.8
& 41.0 & 61.5 & 30.7
& 35.9 & 52.4 & 27.6
& 32.6 & 59.8 & 19.4
& 52.3 & 52.2 & 52.4
& 29.7 & 66.6 & 10.9 \\
& LTC & CLIP
& 54.6 & 19.3 & 72.3
& 60.0 & 73.5 & 32.8
& 45.9 & 67.9 & 35.0
& 42.5 & 51.7 & 37.8
& 49.3 & 74.0 & 37.4
& 58.9 & 59.5 & 58.5
& 37.6 & 72.3 & 36.7 \\
\bottomrule
\end{tabular}
}
\end{table*}

\section{Implementation Details}
\label{ID}
\subsection{Dataset Details}
As outlined in Table~\ref{tab:dataset}, our method is evaluated across multiple benchmarks. Following the protocol established in prior OCD works~\cite{SMILE,PHE}, the categories within each dataset are divided into subsets of seen and unseen categories. Specifically, 50\% of the samples from the seen categories are used to form the labeled training set $\mathcal{D}_S$, while the remainder forms the unlabeled set $\mathcal{D}_Q$ for on-the-fly testing.
\begin{table}[H]
  \setlength{\tabcolsep}{1pt}
  \small
  \renewcommand{\arraystretch}{1}
  \caption{Statistics of datasets used in our experiments.}
  \label{tab:dataset}
  \centering
  \begin{tabular}{lccccccc}
    \toprule
    & CUB & Scars & Pets & Food & CIFAR10 & CIFAR100 & ImageNet100 \\
    \midrule
    $|Y_Q|$ & 200 & 196 & 38 & 101 & 10 & 100 & 100 \\
    $|Y_S|$ & 100 & 98 & 19 & 51 & 5 & 50 & 50 \\
    \midrule
    $|\mathcal D_S|$ & 1.5K & 2.0K & 0.9K & 19.1K & 12.5K & 12.5K & 32.5K \\
    $|\mathcal D_Q|$ & 4.5K & 6.1K & 2.7K & 56.6K & 37.5K & 37.5K & 97.5K \\
    \bottomrule
  \end{tabular}
\end{table}
\subsection{Training Details}

We use the AdamW optimizer with a learning rate of $1 \times 10^{-2}$ for both the backbone and projection head, and apply a weight decay of 0.05. All models are trained for 100 epochs with a consistent batch size of 128 across all datasets to ensure fair comparison with prior methods. For MKEE, we set $\varepsilon = 0.05, \sigma_{0} = 1, \lambda_{\rho} = 0.1$. The warm-up period for MKEE is set to 1 epoch, meaning the perturbation is activated starting from the second epoch. All experiments were run on NVIDIA RTX 4090 GPUs.

\label{sub:training}

\subsection{Compared Methods Details}
Following the experimental setup in SMILE~\cite{SMILE} and PHE~\cite{PHE}, we evaluate our method against several competitive baselines. Below we provide brief descriptions of each:

\noindent \textbf{Ranking Statistics (RankStat).}~\cite{RankStat} AutoNovel employs ranking-based heuristics, using the top-3 indices from feature embeddings to encode each category. This approach is naturally compatible with OCD due to its lightweight descriptor and nonparametric clustering behavior. To ensure fairness, we use the same DINO-ViT-B-16 backbone and discard auxiliary training stages that require access to unlabeled unknowns. The embedding is projected to 32 dimensions, yielding a total prediction space of $C_{32}^3 = 4,906$, which is on par with SMILE ($2^{12} = 4096$) when using 12-bit binary codes.

\noindent \textbf{Winner-Take-All (WTA).}~\cite{WTAjia2021joint} As an alternative to RankStat, WTA mitigates reliance on global embedding order by selecting index maxima within local feature groups. We divide the 48-dimension embedding into three parts and extract the index of the maximum value in each, forming a descriptor of length 3. The resulting prediction space is $16^3 = 4096$, aligning with other methods for consistent comparison.

\noindent \textbf{Sequential Leader Clustering (SLC).}~\cite{hartigan1975clustering} SLC is a classic online clustering method for streaming data. We train the encoder using only labeled support data and apply SLC on extracted features during test time. Hyperparameters are tuned on CUB and fixed across all datasets to ensure comparability.

\noindent \textbf{Meta-Learning for Domain Generalization (MLDG).}~\cite{MLDGli2018learning} Unlike conventional NCD approaches that leverage both support and query sets, OCD restricts training to known-class samples, posing a generalization challenge. To address this, we adapt MLDG for the OCD setting, treating different classes as meta-train and meta-test domains in each iteration. MLDG is then applied on top of our baseline model to promote domain-robust feature learning across class boundaries.

\section{Additional Results under Varying Hyperparameters}
\label{ARU}
\subsection{Effectiveness of AutoTau}
\begin{figure*}[t]
  \centering
  \includegraphics[width=1.0\textwidth]{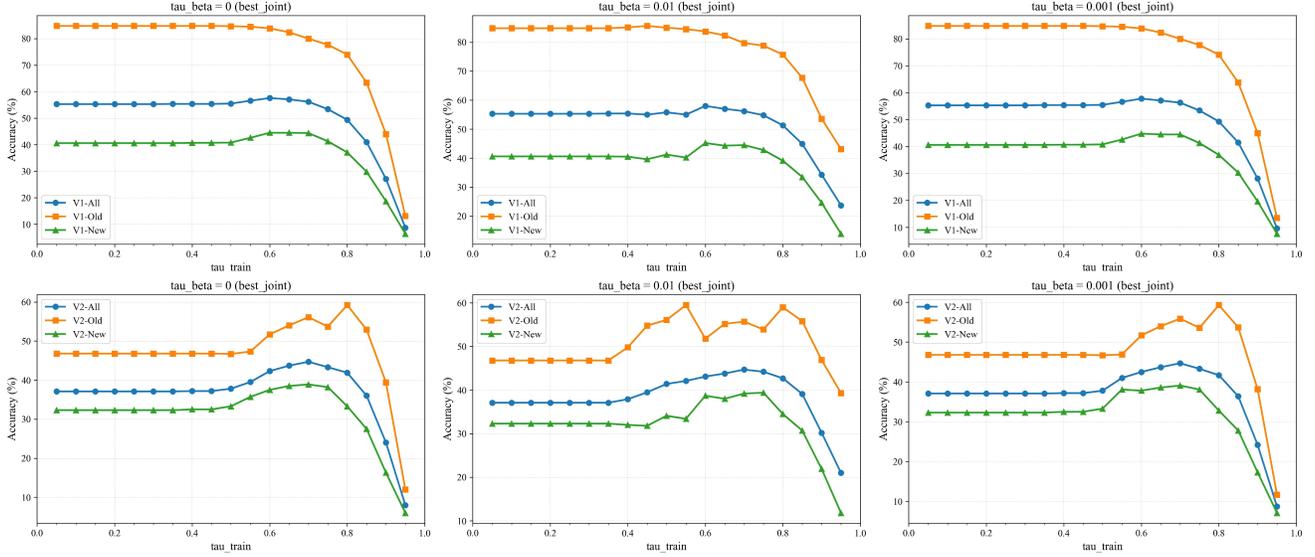}
  \caption{Performance comparison of AutoTau against fixed $\tau_{\text{init}}$ values using the Strict Hungarian and Greedy Hungarian metrics. AutoTau improves model performance by mitigating the impact of suboptimal initial thresholds.}
  \label{fig:suptau}
  \vspace{-2ex}
\end{figure*}

\begin{figure}[t]
    \centering
    \includegraphics[width=1\linewidth]{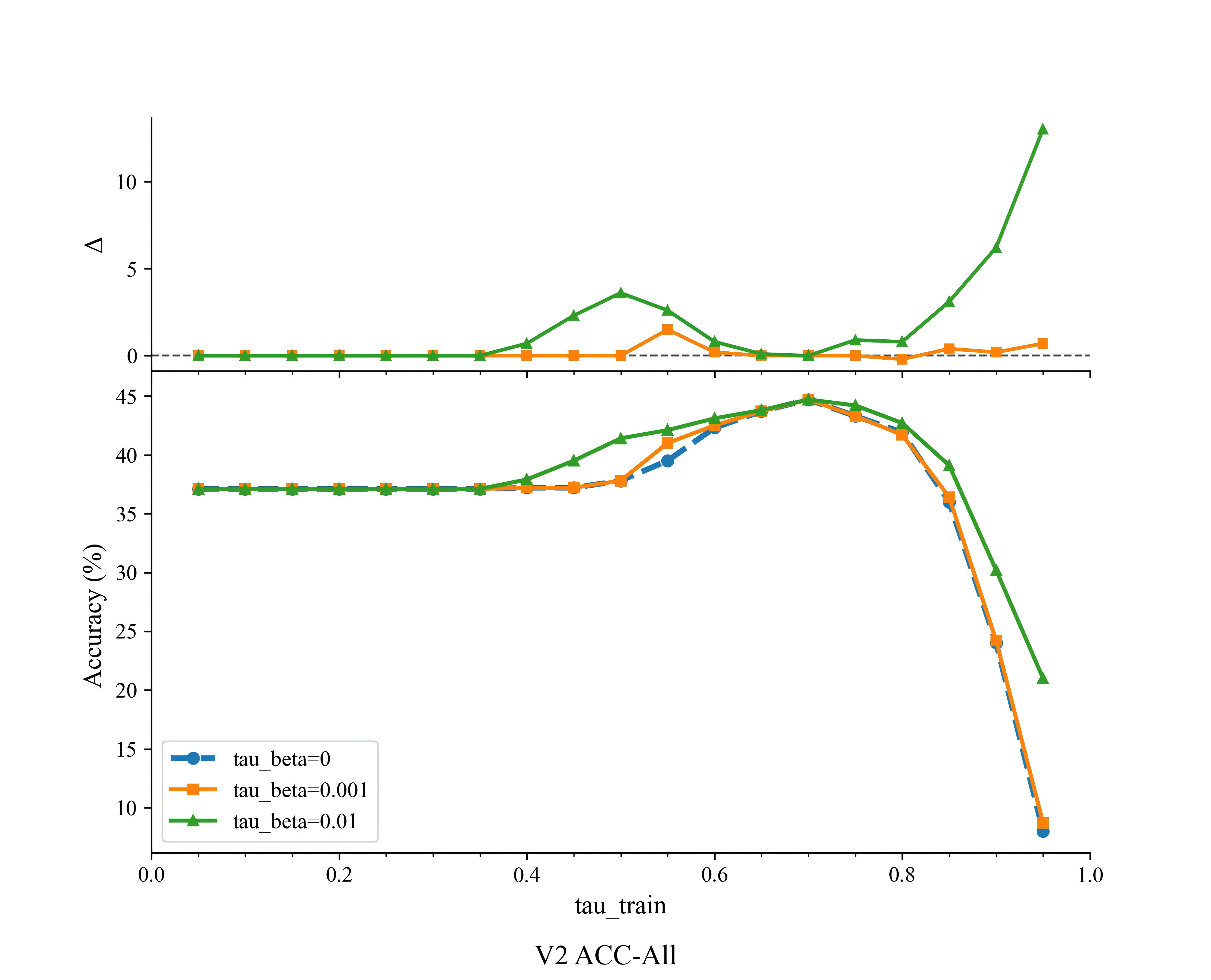}
    \caption{Impact of AutoTau on performance with different $\tau_{\text{init}}$ values. The improvement is more significant in the Strict Hungarian metric, reaching up to 5\% for suboptimal $\tau_{\text{init}}$ values.}
    \label{fig:suptau2}
    \vspace{-2ex}
\end{figure}

We further investigated the effectiveness of the AutoTau mechanism on the CUB dataset. Using both the Strict Hungarian and Greedy Hungarian metrics, as shown in Fig.~\ref{fig:suptau}, AutoTau significantly mitigates the impact of suboptimal initial thresholds ($\tau_{\text{init}}$) and improves model performance. Notably, the enhancement is more pronounced when evaluated using the Strict Hungarian metric. As demonstrated in Fig.~\ref{fig:suptau2}, the improvement reaches around 5\% for certain suboptimal initial values of $\tau_{\text{init}}$.

\subsection{Additional Hyperparameter Sensitivity}

\begin{figure}[t]
    \centering
    \includegraphics[width=0.8\linewidth]{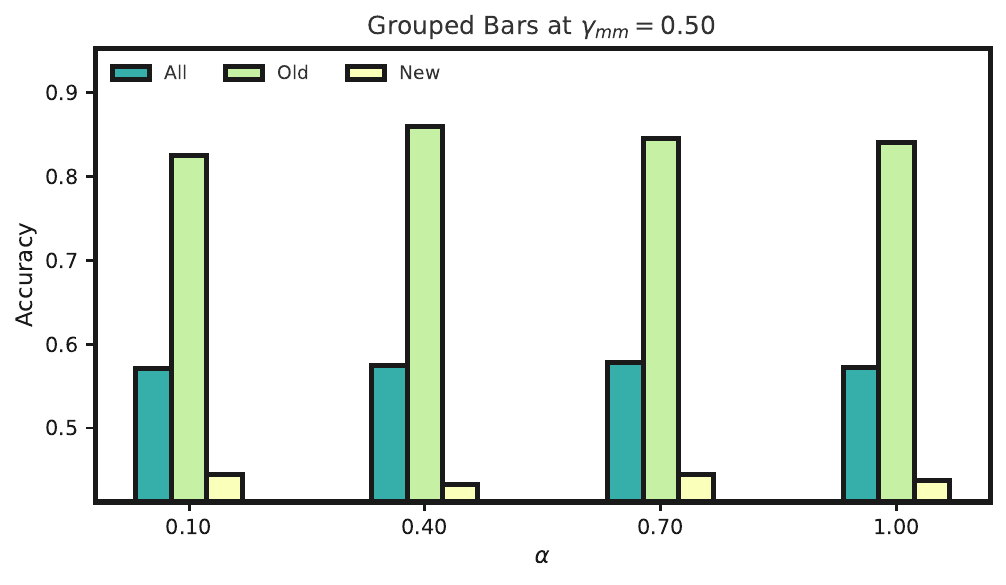}
    \includegraphics[width=0.8\linewidth]{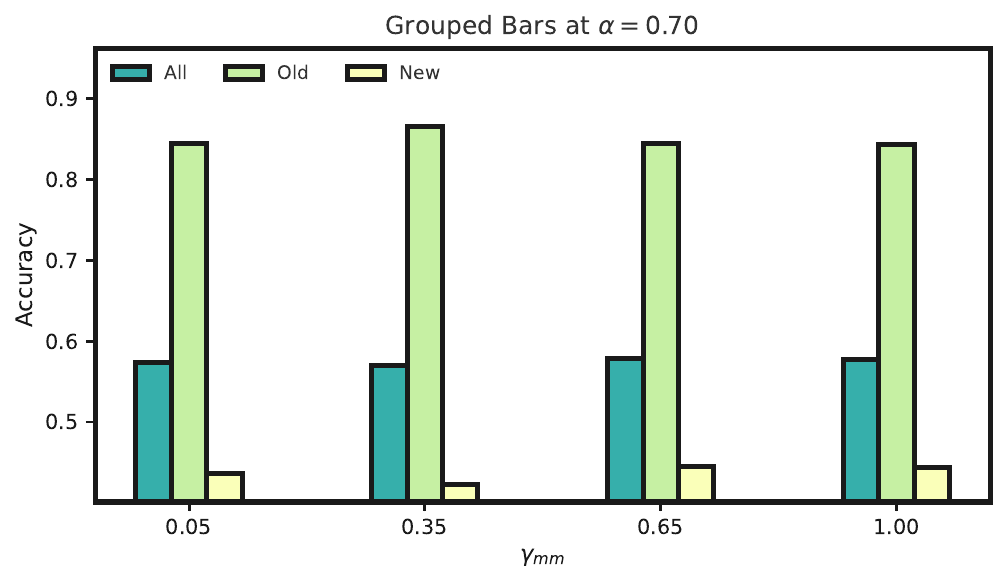}
    \caption{Parameter sensitivity analysis for $\alpha$ and $\gamma_{mm}$. The results show the performance variations with different values of these hyperparameters.}
    \label{fig:supploss}
    \vspace{-2ex}
\end{figure}

To assess the sensitivity of the model to hyperparameters, we conducted experiments with varying values of $\alpha$ and $\gamma_{mm}$. Specifically, we performed parameter sensitivity analysis with $\alpha=0.7$ and $\gamma_{mm}=0.5$. The results, as shown in Fig.~\ref{fig:supploss}, demonstrate that selecting the appropriate parameters can influence the model's performance by approximately 2\%. 

\section{Theoretical Analysis of MKEE}
\label{MKEEEEE}

This appendix analyzes why MKEE (\emph{Minimizing Kernel Energy \& Maximizing Entropy}) improves novel-category discovery while slightly lowering known-class accuracy in On-the-Fly Category Discovery (OCD). The analysis  explain the trade-offs observed in experiments, such as the significant gains in novel class accuracy accompanied by a minor decline in known class performance. 

\subsection{Rationale for One-Step Gradient Approximation Perturbation}

The objective of MKEE is to generate pseudo-unknown samples $x_{\text{pus}}$ by maximizing entropy and minimizing kernel density to encourage model uncertainty:
$$
\mathcal{J}(x) = H(p(y|x)) - \lambda_1 \rho(x),
$$
where $H(p(y|x)) = -\sum_{c} p_c \log p_c$ is the predictive entropy, and $\rho(x)$ estimates the kernel density of the sample in the known feature space. The Mixup sample $x_{\text{mix}} = \lambda x_i + (1-\lambda) x_j$ is perturbed via gradient ascent:
$$
x_{\text{pus}} = x_{\text{mix}} + \varepsilon \cdot \frac{\nabla_x \mathcal{J}(x)}{\|\nabla_x \mathcal{J}(x)\|_2}.
$$
These samples are used in training through a max-margin loss (as described in Section 3.3 of the main text) to jointly optimize known and unknown categories. To justify the one-step rule, we apply Taylor’s theorem with Lagrange remainder: for any small $\Delta x$,
$$
\mathcal{J}(x + \Delta x) = \mathcal{J}(x) + \nabla_x\mathcal{J}(x)^\top \Delta x + \tfrac{1}{2} \Delta x^\top H_x(\xi) \Delta x,
$$
where $H_x(\xi)$ is the Hessian at some point $\xi$. If $\|H_x(\xi)\|_2 \leq L$ locally, then
$$
\mathcal{J}(x + \Delta x) \geq \mathcal{J}(x) + \nabla_x\mathcal{J}(x)^\top \Delta x - \tfrac{L}{2} \|\Delta x\|_2^2.
$$
Maximizing the right-hand side over the ball $\{\|\Delta x\|_2 \leq \varepsilon\}$ yields the optimal step $\Delta x^\star = \varepsilon \, \nabla_x\mathcal{J}(x) / \|\nabla_x\mathcal{J}(x)\|_2$, with an improvement lower bound:
$$
\mathcal{J}(x + \Delta x^\star) \geq \mathcal{J}(x) + \varepsilon \|\nabla_x\mathcal{J}(x)\|_2 - \tfrac{L}{2} \varepsilon^2.
$$
This derivation validates the practicality of the one-step gradient rule, as it ensures a guaranteed increase in $\mathcal{J}(x)$ under reasonable smoothness assumptions.

\subsection{Why MKEE Slightly Reduces Known-Class Accuracy?}

\noindent\textbf{Impact of Entropy Maximization on Known Classes.} Consider a classification model that outputs a probability distribution $p(y|x)$. For known classes, the training objective minimizes the cross-entropy loss to encourage high-confidence predictions (i.e., entropy minimization):
$$
\mathcal{L}_{\text{ce}} = -\mathbb{E}_{x \sim \mathcal{D}_{\text{known}}} \left[ \sum_{c=1}^{K} y_c \log p_c(x) \right],
$$
where $y_c$ is the one-hot encoded label. This pushes the model toward Dirac-like distributions for known samples, resulting in low entropy. In contrast, MKEE generates pseudo-unknown samples that require high entropy (low confidence). This conflict introduces an adversarial dynamic during optimization.

For a linear classifier with logits $\ell(x) = Wx + b$ and softmax probabilities $p_c(x) = \exp(\ell_c(x)) / \sum_k \exp(\ell_k(x))$, the predictive entropy $H(p(y|x))$ is a function of $\ell(x)$. Minimizing $\mathcal{L}_{\text{ce}}$ drives $\ell(x_{\text{known}})$ to extreme values, reducing entropy, while entropy maximization for $x_{\text{pus}}$ pulls $\ell(x_{\text{pus}})$ toward zero to equalize logits. When both objectives are combined, the total loss effectively includes an entropy regularization term:
$$
\mathcal{L}_{\text{total}} = \mathcal{L}_{\text{ce}} + \beta \mathbb{E}_{x \sim \text{pseudo-unknown}} \left[ H(p(y|x)) \right].
$$
This regularization softens decision boundaries, improving generalization to novel classes but reducing the sharpness of predictions for known classes, leading to a slight accuracy drop. Notably, MKEE achieves this indirectly through sample generation rather than direct loss modification, which avoids training instability associated with opposing loss terms.

\noindent\textbf{Effect of Kernel Density Minimization.} The kernel density term $\rho(x)$ in MKEE pushes samples away from dense regions of the known class distribution. While this exposes the model to out-of-distribution samples, it is beneficial for novel class discovery, and also biases decision boundaries away from known class centroids. Statistically, the density minimization acts as a penalty term:
$$
\mathcal{L}_{\text{total}} \approx \mathcal{L}_{\text{ce}} + \beta_1 H(p(y|x_{\text{pus}})) - \beta_2 \rho(x_{\text{pus}}),
$$
promoting robustness at the cost of marginal accuracy for known classes near decision boundaries. This explains the observed trade-off where known class accuracy decreases slightly while novel class accuracy improves substantially.

\subsection{Correspondence with Experiments}

Empirical results align with the theoretical analysis: MKEE consistently enhances novel class accuracy with a minor known class degradation (e.g., -1\% on datasets like CIFAR-10). This trade-off is acceptable in OCD, where novel class discovery is prioritized. The adaptive threshold calibration in MKEE (Section 3.3 of the main text) mitigates this decline by dynamically balancing the novelty trigger. The parameters $\lambda_1$ and $\beta$ allow tuning the trade-off, though future work could explore finer sample generation strategies to minimize known class impact while preserving novel gains.

\section{Limitations Discussion}
\label{limit}

\noindent \textbf{Balance Between New and Old Category Accuracy.} In our experiments, we observed that the accuracy of the old categories may slightly degrade when using the MKEE approach. This is a trade-off we face in the pursuit of significantly improved new category discovery capabilities. While the proposed method excels at detecting new classes, this comes at the expense of some loss in performance on the old categories. In future work, a key challenge will be to develop methods that mitigate the forgetting of old category knowledge while maintaining the ability to effectively discover new categories. Techniques such as continual learning, knowledge distillation, or regularization strategies might be useful in addressing this limitation and ensuring more balanced performance across both known and novel categories.

\noindent \textbf{Exploring the Integration of Textual Modality.} Our current approach leverages the CLIP backbone, which primarily utilizes visual data. However, incorporating additional modalities, particularly textual information, into the OCD task presents an intriguing avenue for future exploration. The integration of text can potentially enhance the model's ability to generalize across categories and improve the discovery of novel classes, especially in scenarios where labeled data is sparse or ambiguous. Exploring how to effectively combine multimodal data, such as by aligning text and image representations, could further enrich the model's understanding of complex class relationships.

\end{document}